# A Comprehensive Analysis on Machine Learning based Methods for Lung Cancer Level Classification

Shayli Farshchiha[a], Salman Asoudeh[b], Maryam Shavali Kuhshuri[c], Mehrshad Eisaei[d] ,Mohamadreza Azadi[e] ,Saba Hesaraki[f]

[a] Isfahan University of Technology, Department of Electrical and Computer Engineering, Faculty of Engineering, Isfahan, Iran.

[b] Faculty of Computer, College of Technical and Engineering, University of Velayat, Iran.

[c] University of Isfahan Department of Electrical Engineering, Faculty of Engineering, University of Isfahan, Isfahan, Iran.

[d] University of Mazandaran, Mazandaran, Iran.

[e] Department of Computer Science,Shamsipour technical college Tehran, Iran.

[f] Islamic Azad University Science and Research Branch, Faculty of Mechanics, Electrical Power and Computer

## Abstract

Lung cancer is a major issue in worldwide public health, requiring early diagnosis using stable techniques. This work begins a thorough investigation of the use of machine learning (ML) methods for precise classification of lung cancer stages. A cautious analysis is performed to overcome overfitting issues in model performance, taking into account minimum child weight and learning rate. A set of machine learning (ML) models including XGBoost (XGB), LGBM, Adaboost, Logistic Regression (LR), Decision Tree (DT), Random Forest (RF), CatBoost, and k-Nearest Neighbor (k-NN) are run methodically and contrasted. Furthermore, the correlation between features and targets is examined using the deep neural network (DNN) model and thus their capability in detecting complex patterns is established. It is argued that several ML models can be capable of classifying lung cancer stages with great accuracy. In spite of the complexity of DNN architectures, traditional ML models like XGBoost, LGBM, and Logistic Regression excel with superior performance. The models perform better than the others in lung cancer prediction on the complete set of comparative metrics like accuracy, precision, recall, and F-1 score.

## 1. Introduction

Lung cancer is a major global health concern, with the unfortunate distinction of being the most common cause of cancer-related deaths and one of the most diagnosed cancers in the world. Approximately 2.20 million individuals are diagnosed each year with the debilitating lung cancer diagnosis, a staggering 75% of whom die within five years. The intricate inter-tumor heterogeneity and drug resistance make it extremely difficult to effectively treat cancer. Recent advances in cancer research techniques have brought with them a new era of cooperative work, leading to the establishment of large databases of clinical, imaging, and sequencing information. These data provide researchers with broad avenues of studying the intricate dynamics of lung cancer, including diagnosis and treatment data and clinical outcomes. The availability of -omics research, including genomics, transcriptomics, proteomics, and metabolomics, has greatly expanded the repertoire of tools for investigation. However, the incorporation of different types of high-dimensional data into clinical practice is extremely time-consuming and demands a lot of expertise. This emphasizes the increasing role of machine learning (ML) models, which can automatically detect inherent patterns in data to assist health care professionals in their decision-making. In spite of their potential, there is hesitation on the part of researchers and practitioners in investing in decision-making with expensive datasets. ML is a breath of fresh air in that it enables effective decision-making, accuracy, cost-effectiveness, and reproducibility regardless of geographical or human resource limitations. In this research, we aim to utilize machine learning algorithms to create efficient predictive models for the classification of lung cancer types, based on critical protein characteristics. By integrating feature selection, decision tree induction, and clustering methods, we intend to enhance the predictive accuracy of the models, thereby facilitating improvements in lung cancer diagnosis and treatment. Moreover, we are confronted with the challenges of clinical and in-situ patient surveillance, with the aim of delivering economically viable and time-efficient alternatives.

Machine learning (ML), a branch of artificial intelligence (AI), has been a critical tool for cancer phenotyping and treatment for many years. In this research, we employ a broad spectrum of ML models, ranging from classic to modern algorithms, to offer comprehensive information to patients, researchers, and experts. Additionally, we scrutinize often underestimated machine learning model hyperparameters, such as minimum child weights and learning rates, to show their phenomenal impact on accuracy errors through thorough analysis. With the increasing sophistication of machine learning techniques in the diagnosis and treatment of lung cancer, this study aims to maximize patient outcomes and streamline decision-making in clinical practice.

## Related Works

Machine learning has emerged as the backbone of modern medicine, with enormous potential for the enhancement of the accuracy of disease diagnoses and prognoses like lung cancer. Using machine learning models, scientists are not only explaining disease mechanisms but also revolutionizing clinical practice by adopting personalized treatment strategies and predictive model methods. In the case of lung cancer, life expectancy after surgery estimation is most important in treatment planning and patient counseling. The use of data mining prediction models like Decision Tree, Naive Bayes, and Artificial Neural Network makes personalized treatment planning by clinicians based on the individual features of every patient possible [1]. Additionally, the use of stratified 10-fold cross-validation analyses also ensures the validity and relevance of these predictive models, further making them more valuable in clinical application. Moreover, the application of classification algorithms for diagnosing brain tumors offers useful information for diagnosing lung cancer as well. By evaluating various classifiers' performance based on volumetric and locational features, scientists gain critical information on the algorithms' ability to distinguish between classes [2]. Such information not only facilitates early treatment and detection but also helps clinicians to characterize tumors and plan treatment accordingly.

In the area of lung cancer diagnosis, the excellent accuracy displayed by classifiers like Support Vector Machine says a lot about the transformative power of machine learning in medical practice [3]. Using various algorithms like KNN, SVM, NN, and Logistic Regression, doctors are capable of enhancing their decision-making capabilities related to diagnosis, staging, and treatment plans. In addition, the intersection of diagnostic devices with machine learning in healthcare indicates promising improvement in patient outcomes and efficient resource utilization in healthcare organizations. Additionally, studies on segmentation algorithms such as Naïve Bayes and the Hidden Markov Model are an important contribution to the accuracy of lung tumor detection [4].

By outlining tumor margins and describing tissue morphology, such algorithms allow clinicians to detect minute abnormalities and distinguish between benign and malignant lesions [5–8]. In addition, the creation of algorithmic flowcharts for the detection of brain tumors highlights the interdisciplinarity of oncological diagnostic research with implications for lung cancer diagnosis and treatment [9]. The integration of machine learning algorithms into practice is a paradigm shift in the diagnosis of oncology and personalized medicine strategies. By leveraging the potential offered by evidence based on data, health care providers can improve patient outcomes, optimize resource use, and, in the end, our shared knowledge of lung cancer and its treatment. As research in this area continues, the ability of machine learning to transform the provision of health care is limitless.

## 2. Materials and Methods

Prediction of lung cancer using machine learning is extremely difficult due to the fact that it needs collecting and organizing datasets from various regions across the world.

This is done to make datasets balanced and representative of various locations and scenarios. To surpass this, detection and classification levels of lung cancer data are collected from publicly available datasets published by reputable agencies such as the World Health Organization (WHO), Kaggle, and Google datasets. Consolidating data from various sources, we aim to create an approximately balanced dataset.

### 3.1 Dataset

The initial step is the gathering of data from various publicly available datasets provided by trusted organizations like WHO, Kaggle, and Google datasets. The data gathered contain numerous attributes like patient ID, age, gender, and various environmental and lifestyle factors known to affect the risk of lung cancer. The target variable is the lung cancer level, which has three stages: low, medium, and high.

.

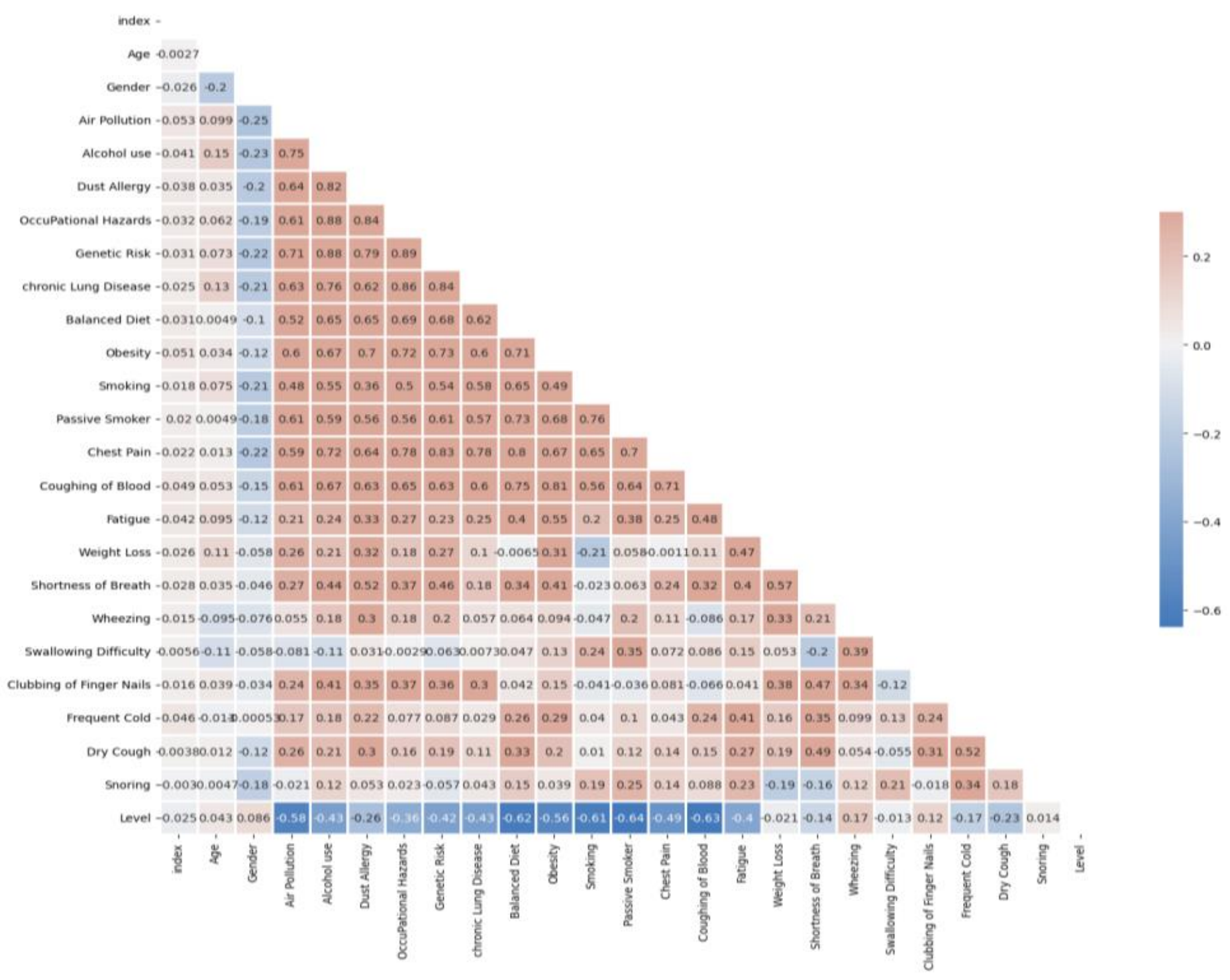

**Figure 1.** Lung Cancer Correlation.

## 3.2 Preprocessing

Once the datasets are collected, they undergo a cleaning and preprocessing phase with the objective of establishing meaningful relationships between the various attributes. The features are analyzed for correlations to acquire information on how they influence the target variable. Correlation plots are created to visualize the relationships among the features. Features that have correlations greater than a certain value (for instance, 0.5) are deemed good for merging, thus creating new columns in the dataset. The method increases the efficiency of the features in lung cancer level prediction.

## 3.3 Correlation Analysis

Correlation plots are used to examine relationships between different features and the target variable. Strongly positive correlating features (e.g., those greater than 0.5) are merged to create new columns with increased predictive power. Other features with minimal or negative correlation are under close scrutiny in order to capture the effect of the feature on the target variable.

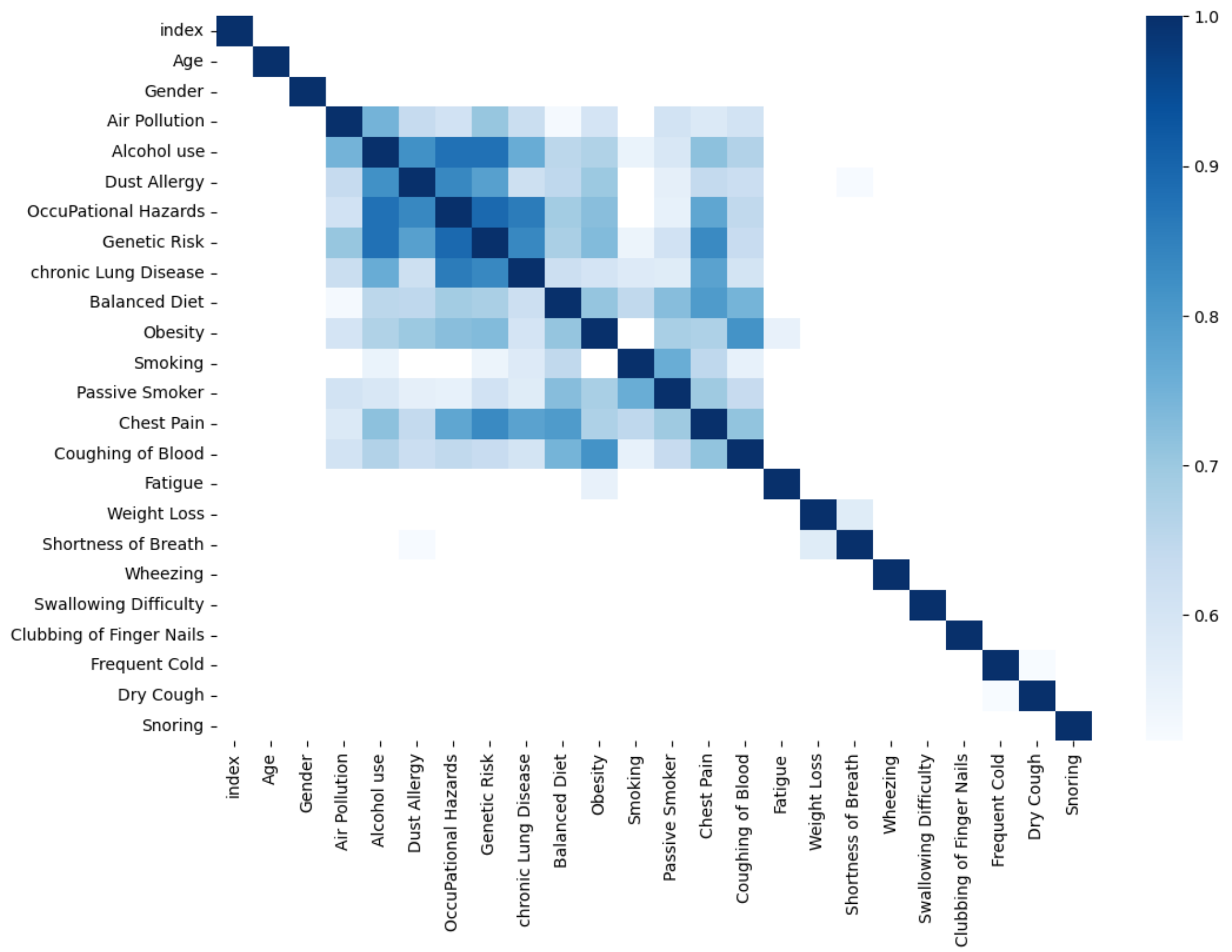

**Figure 2.** Lung Cancer Correlation More than 0.5.

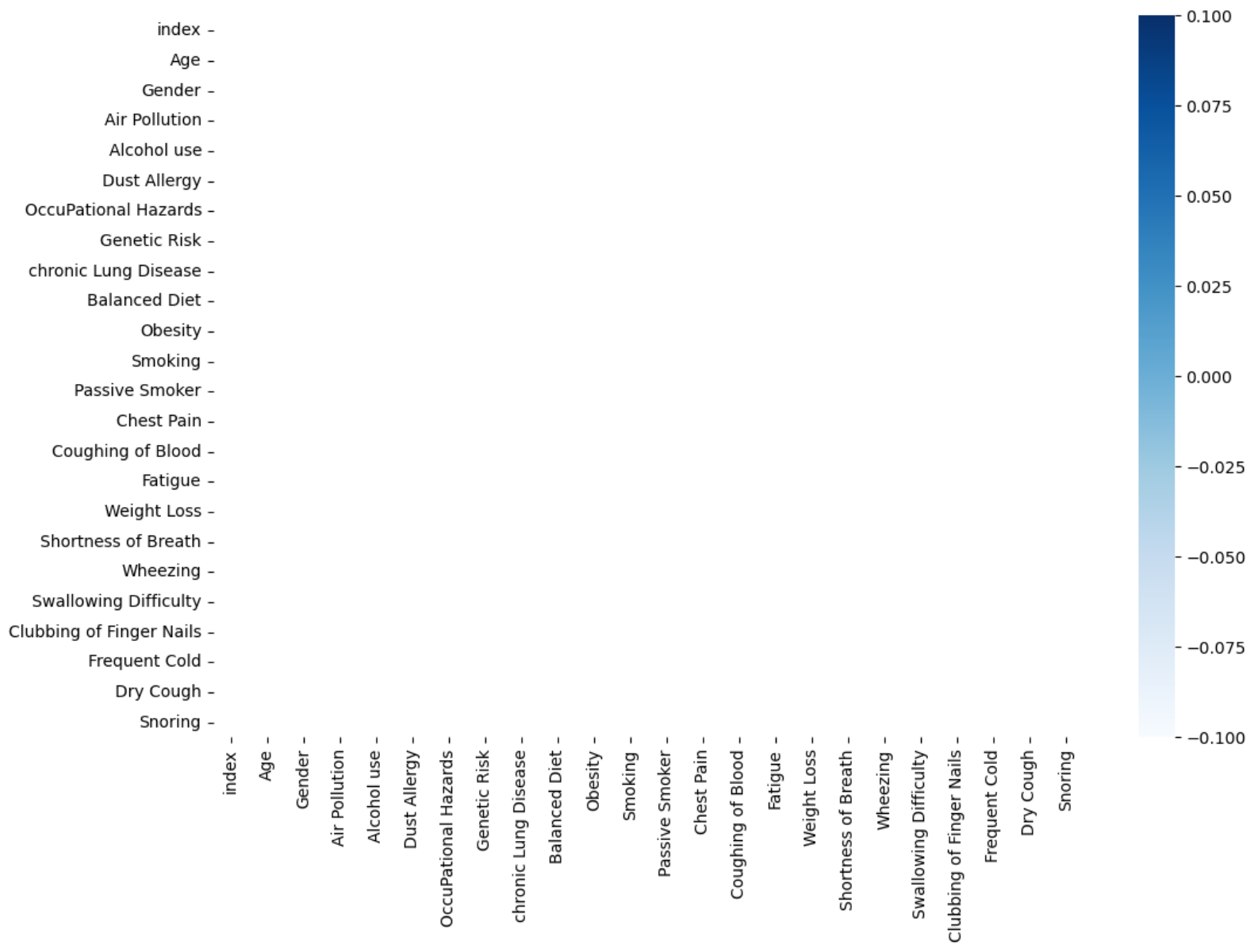

**Figure 3.** Lung Cancer Correlation Less Than 0.4.

## Splitting Data

Following the preprocessing stage, datasets are split into training and testing sets in an attempt to counteract overfitting risk. Features are normalized through the use of MinMaxScaler, and a balanced dataset is attained through the use of the Synthetic Minority Over-sampling Technique (SMOTE). There are 876 samples in the training subset and 219 samples in the testing subset. In addition, k-fold cross-validation with k=5 is performed in an attempt to further reduce overfitting probability.

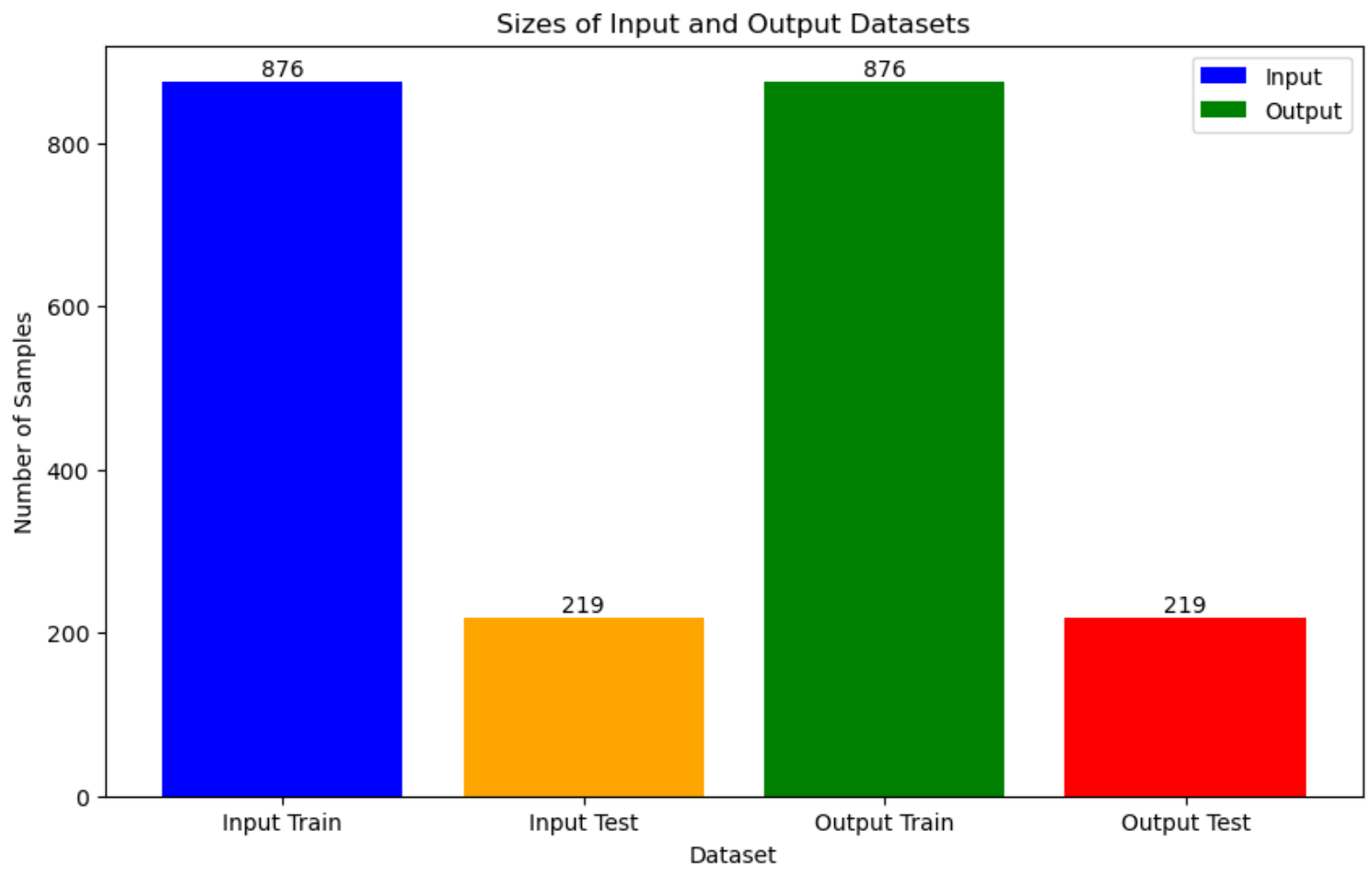

**Figure 4.** Splitting Lung Cancer Dataset.

## ML Models

A total of nine machine learning models are applied for the prediction of lung cancer levels: XGBoost, LGBM, AdaBoost, LogisticRegression, Decision Tree, Random Forest, CatBoost, KNN, and DNN. The models are selected according to their suitability for the task and potential in delivering high accuracy in the prediction of lung cancer. The models are trained with the preprocessed datasets and evaluated with various performance measures in order to determine their efficiency in predicting lung cancer levels.

These models were chosen due to their efficiency in classification tasks and their capability to interpret complicated relationships within the dataset. With a variety of machine learning algorithms, we are hoping to create strong and precise predictive models for the detection and classification of lung cancer.

### Extreme Gradient Boosting (XGBoost)

Extreme Gradient Boosting, better referred to as XGBoost, is an ensemble learning algorithm that is renowned for its scalability and efficiency in computations. Through iteratively training a series of weak learners, XGBoost incrementally improves the model's predictive power, successfully capturing complex patterns in data. Its capability to support both numerical and categorical features, in addition to being resilient to overfitting, renders it particularly well-adapted to classification problems.

### Light Gradient Boosting Machine (LGBM)

Light Gradient Boosting Machine, or LGBM, is a gradient boosting framework specially designed to work effectively with large datasets and high-dimensional features. As opposed to traditional gradient boosting algorithms, LGBM employs a leaf-wise growth approach, which dramatically improves computational efficiency and model performance. Its computation speed, scalability, and capability in dealing with categorical features and imbalanced data render it a top choice for classification problems [1].

### Adaptive Boosting (AdaBoost)

Adaptive Boosting, simply AdaBoost, is an ensemble learning algorithm that sequentially aggregates a collection of weak classifiers to build a stronger classifier. By assigning larger weights to misclassified samples in each pass, AdaBoost prioritizes improving classification accuracy, specifically in regions of feature space that have difficulty with it. With its simplicity, adaptability, and effectiveness in a variety of datasets, it is an important tool for classification tasks [9].

### Logistic Regression

Logistic Regression is a simple linear model that is most commonly used for binary classification tasks. With its simple mechanism, logistic regression proves to have satisfactory performance when a feature and target variable relation is linear, or can at least be approximated with a linear function. With its simplicity, computational effectiveness, and ease of use, logistic regression is most often adopted for creating baseline classification models.

### Decision Trees

Decision Trees are non-parametric supervised learning algorithms that partition feature spaces into hierarchical structures, allowing for simple decision processes. Decision trees have a strong capability for representing complex relations between features and target variable, and therefore, can effectively classify with non-linear decision boundary cases. Decision trees, however, suffer from overfitting [10], especially in cases with deep trees and noisy datasets.

### Random Forest

Random Forest is an ensemble learning algorithm with a base in decision trees, in which many trees are trained individually and then aggregated to produce a prediction. By combining individual tree output, random forests mitigate overfitting and generalize capabilities [11] and have a high tolerance for noise, high-dimensional capabilities, and ease in dealing with high-dimensional datasets, and therefore, can effectively classify in many cases.

### Categorical Boost (CatBoost)

CatBoost constitutes a gradient boosting algorithm specifically designed to effectively work with categorical features. Unlike traditional approaches, CatBoost incorporates a new algorithm for processing categorical variables, therefore, bypassing traditional preprocessing techniques such as one-hot encoding. Overfitting tolerance, efficiency in dealing with big datasets [12], and high predictive accuracy make CatBoost particularly ideal for classification with categorical features.

### K-Nearest Neighbors (KNN)

The K-Nearest Neighbors (KNN) algorithm works as a simple yet powerful instance-based algorithm utilized in classification processes. KNN determines a specific data point's class label through an examination of its k most similar neighbors in feature space and assigning it with the most common class label present in them. Simplicity, adaptability, and its capability to form complex decision borders make KNN a fitting option for a variety of classification scenarios, particularly in cases with non-parametric distributions [13].

### Deep Neural Networks (DNN)

DNNs are a group of neural networks with many layers, which have the ability to learn complex representations directly from raw data [14]. By combining many neurons, DNNs can learn hierarchical features in an unsupervised manner, allowing them to detect complex structures in data. With its ability to manage high-dimensional inputs, learn non-linear relations, and adapt to many datasets, DNNs make them strong tools for classification, specifically in the areas of image, text, and speech processing [15,16].

## Results and Discussion

In our quest to mitigate overfitting and enhance model performance, significant consideration was placed in tracking parameters such as minimum child weight and learning rate in a variety of machine learning (ML) models. In relation to XGBoost model (Figure 5, Figure 6, Figure 7), our efforts produced astounding results, with performance values consistently trending towards 100% accuracy, precision, recall, and F-1 value. This achievement reflects the success of our approach in optimizing model parameters for best performance. With careful hyperparameter tracking, concerns regarding overfitting have been addressed, and both model robustness and generalizability have been guaranteed.

The information gained through confusion matrices (Figure 6) and learning curves provided important feedback about model performance. Notably, our observations showed that an additional incorporation of datasets did not yield significant improvements in model performance, indicative of a saturation point with regard to utility in terms of data. In addition, we discovered that a learning rate exceeding 0.06 produced additional performance improvements with no indications of overfitting and underfitting, and therefore, re-emphasizes the role of parameter optimization in attaining best model performance.

These findings not only contribute to the advancement of ML methodologies but also hold profound implications for clinical practice. By leveraging insights garnered from our analysis, clinicians can make more informed decisions regarding lung cancer diagnosis and treatment, ultimately improving patient care outcomes. Moreover, our approach serves as a testament to the importance of meticulous parameter tuning in maximizing the predictive capabilities of ML models, setting a precedent for future research endeavors in oncological diagnostics and beyond.

.

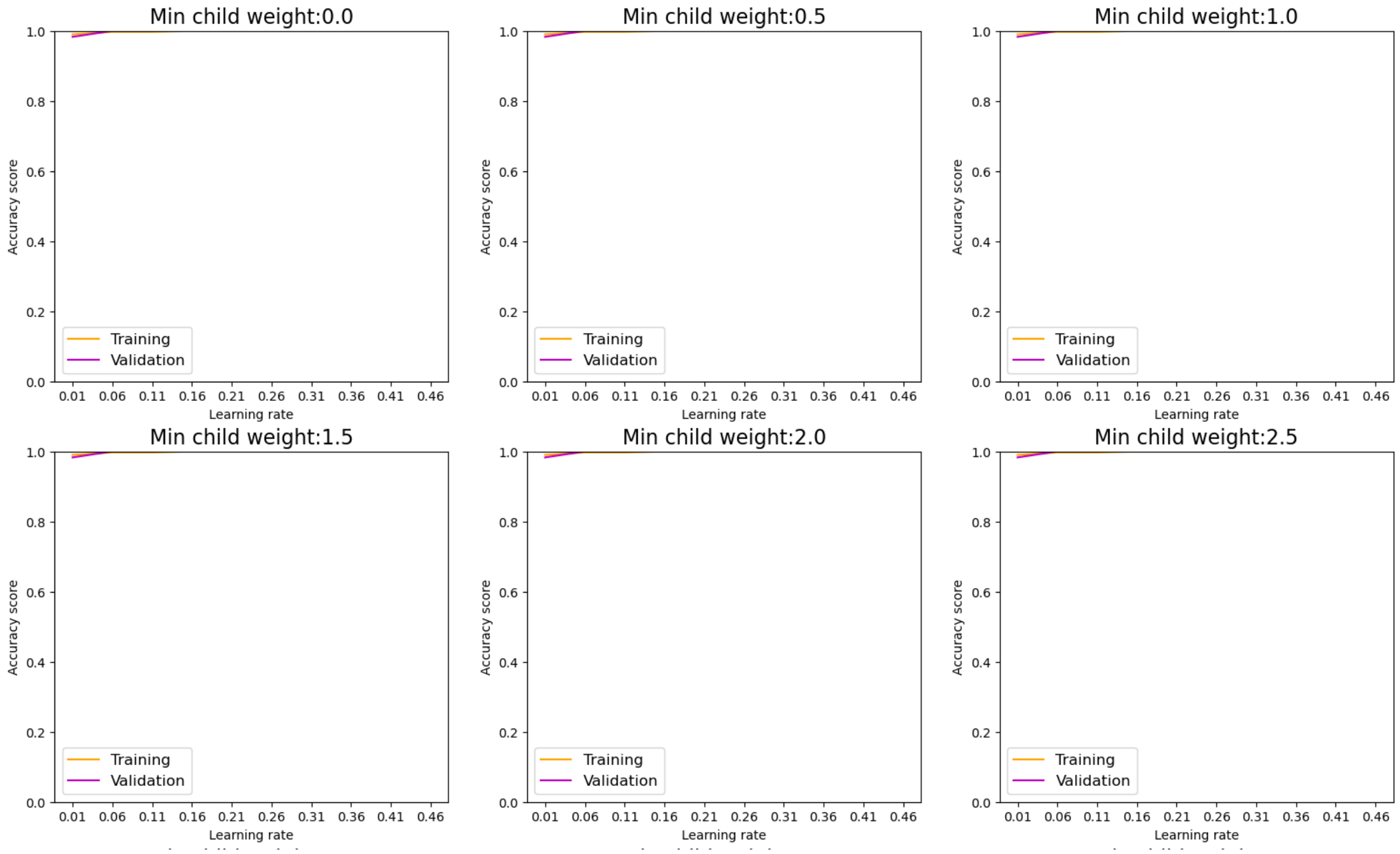

**Figure 5.** Training and Validation plots under consideration of Min child Weight and Learning Rate in XGBoost.

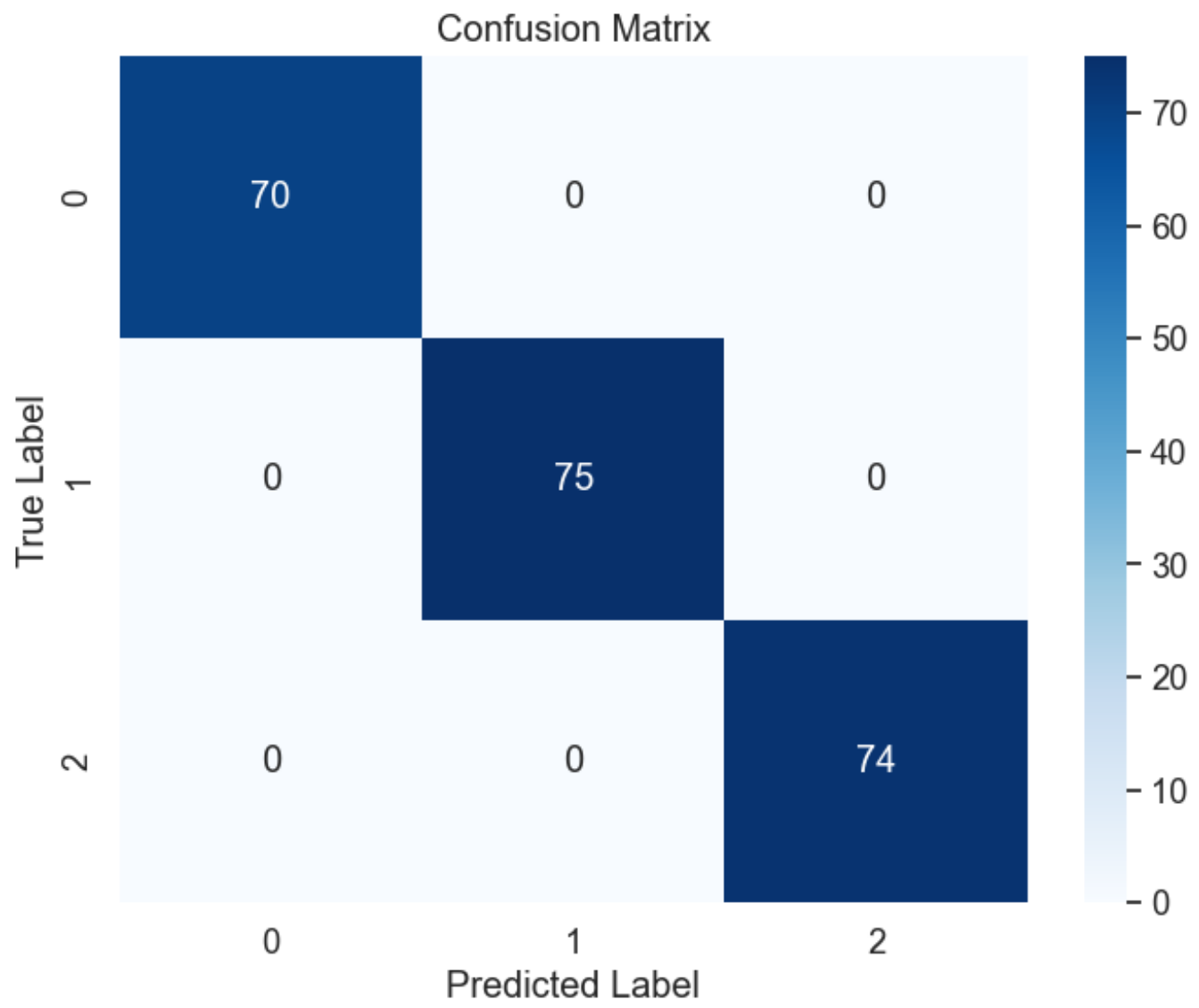

**Figure 6.** Confusion Matrix of XGBoost.

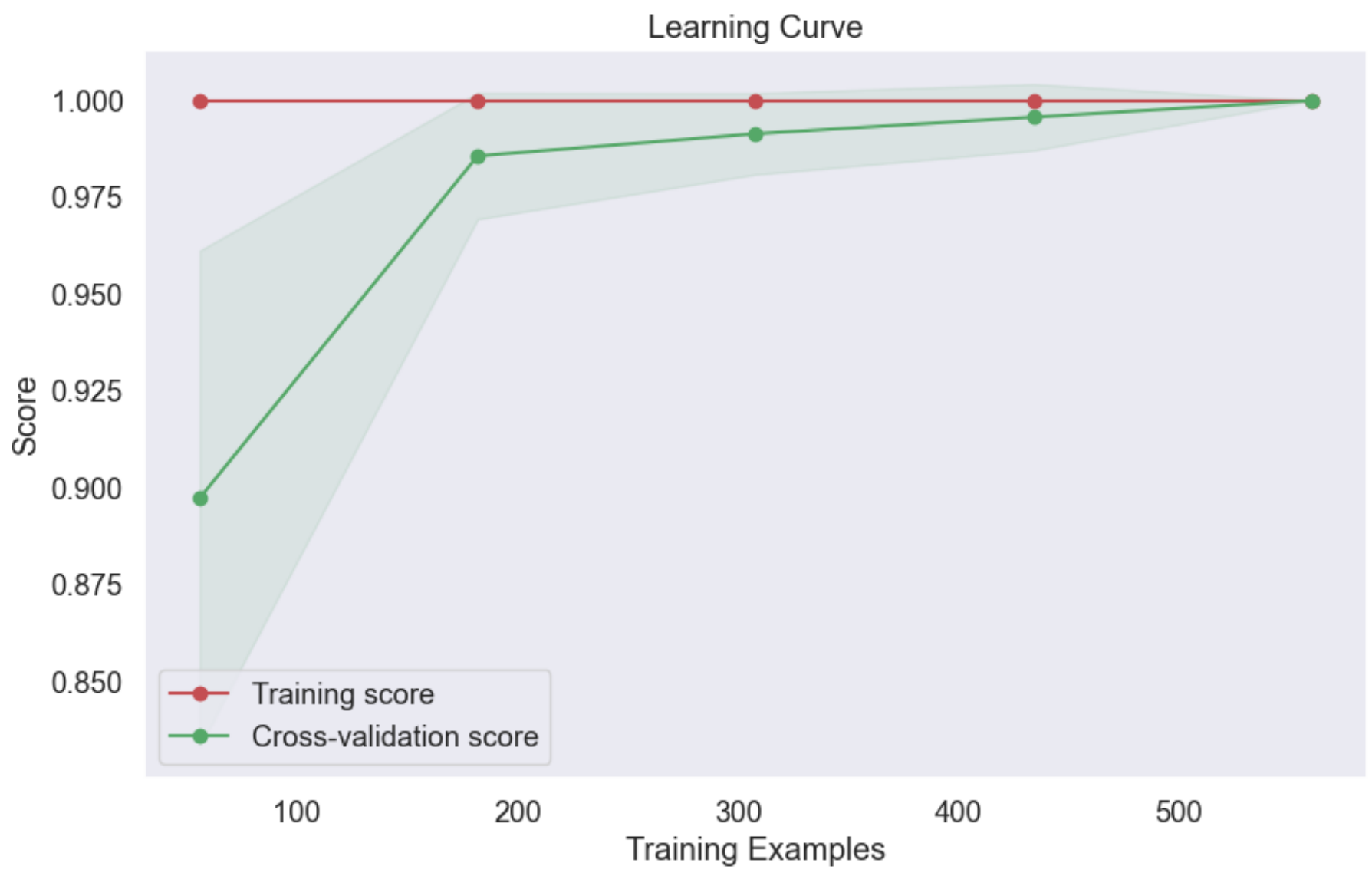

**Figure 7.** Learning Curve of XGBoost.

A comprehensive analysis of hyperparameters for the LGBM model (as seen in Figures 8, 9, and 10) showed uniform trends, confirming our practice of careful model parameter tuning for best performance. An improvement in minimum child weight and learning range (seen in Figure 8) produced a linear improvement in training and validation metrics, inferring a uniform improvement in model parameters. This careful analysis reached performance markers, with LGBM model performance at almost 100% accuracy, precision, recall, and F-1 score (seen in Figure 9). Notably, tests performed during testing confirmed these observations, with demonstration of model robustness and generalizability, and overcoming overfit concerns.

The observations confirm the effectiveness of our approach in a uniform search through model parameter space and improvement in performance. With careful hyperparameter calibration, high performance markers, and demonstration of the importance of parameter tuning in improvement in predictive performance of machine learning algorithms, have been attained. In addition, these observations present important information for clinicians and researchers, with improvements towards specific and reliable diagnostics for lung cancer prognosis and detection.

.

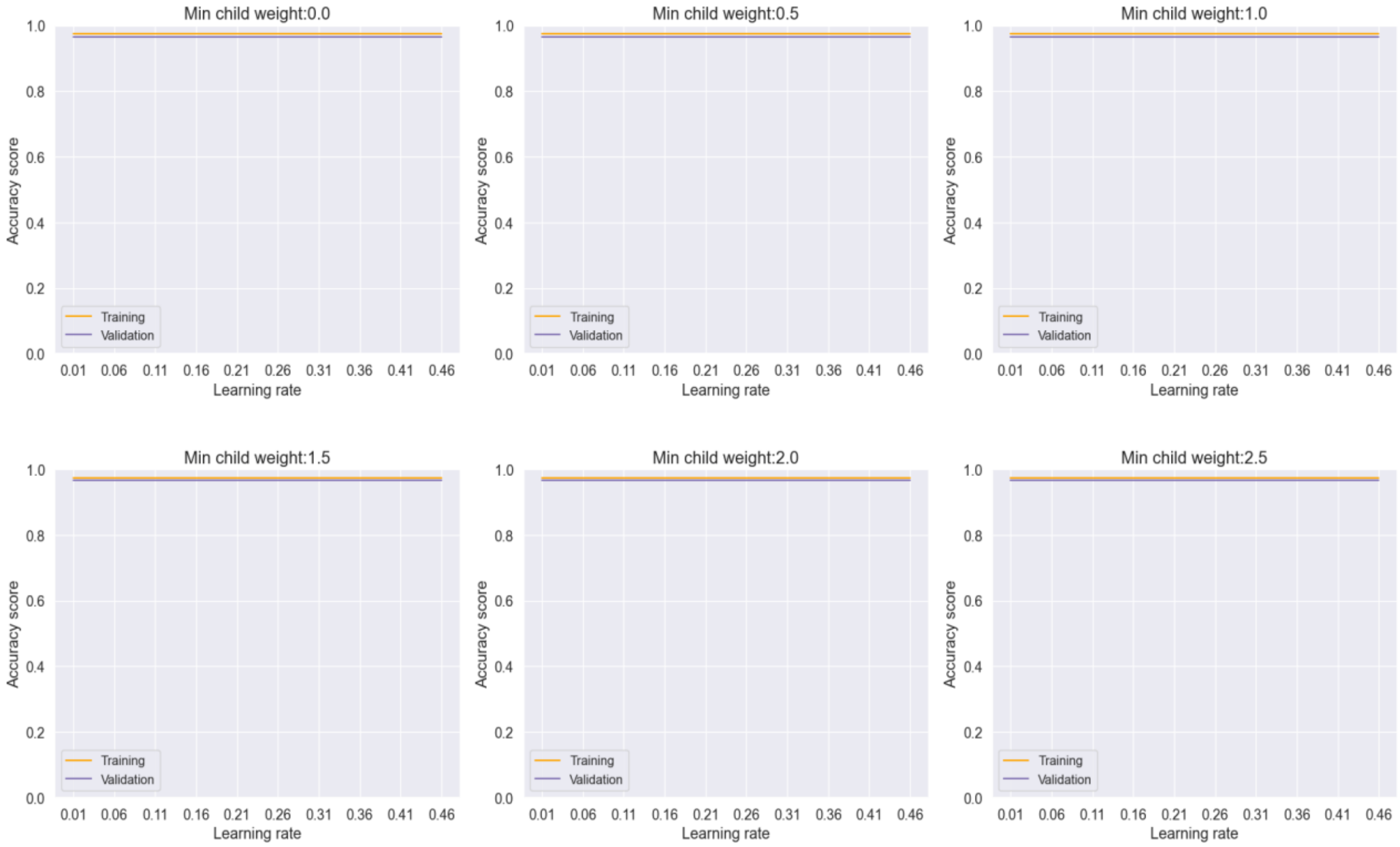

**Figure 8.** Training and Validation plots under consideration of Min child Weight and Learning Rate in LGBM.

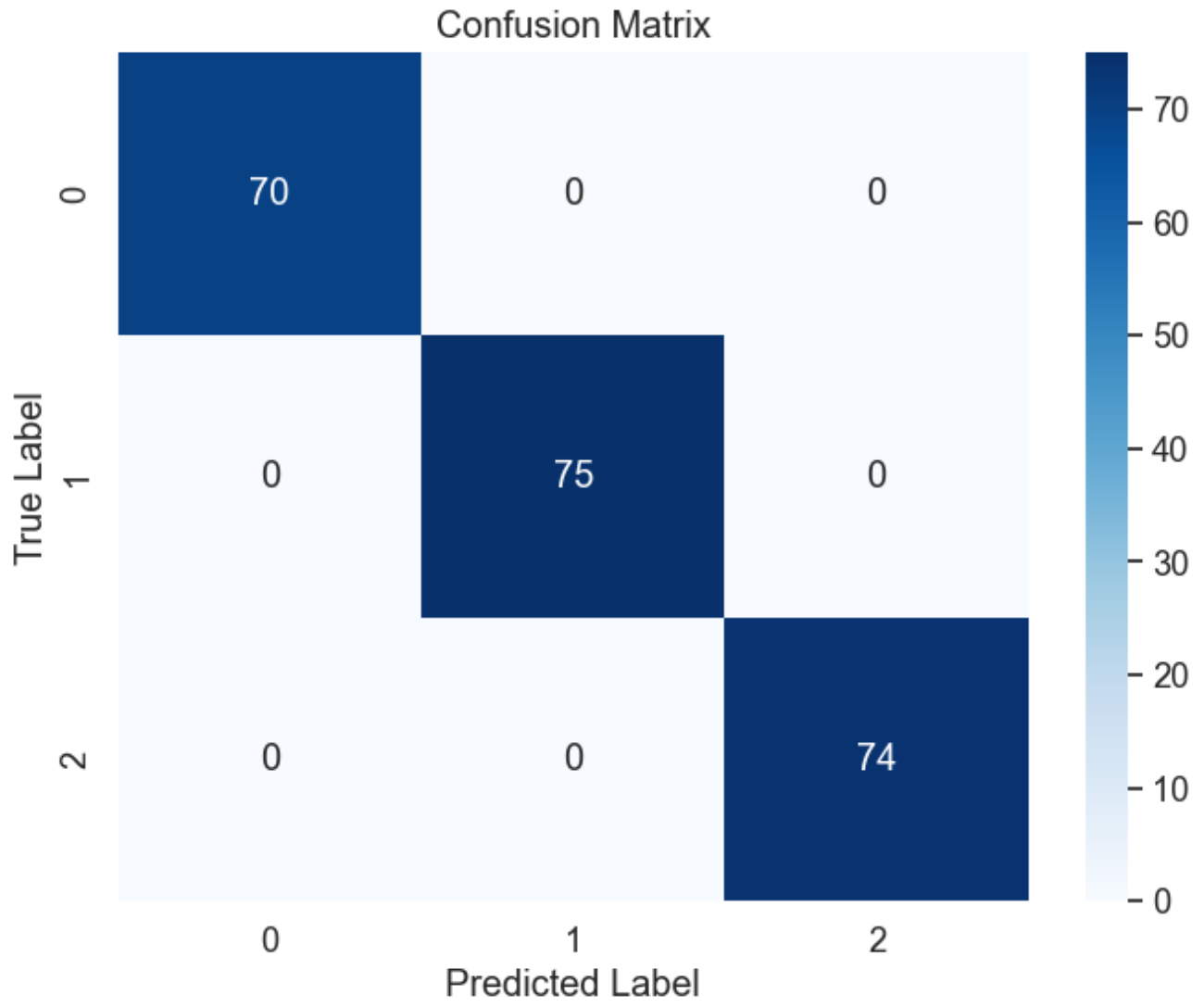

**Figure 9.** Confusion Matrix of LGBM.

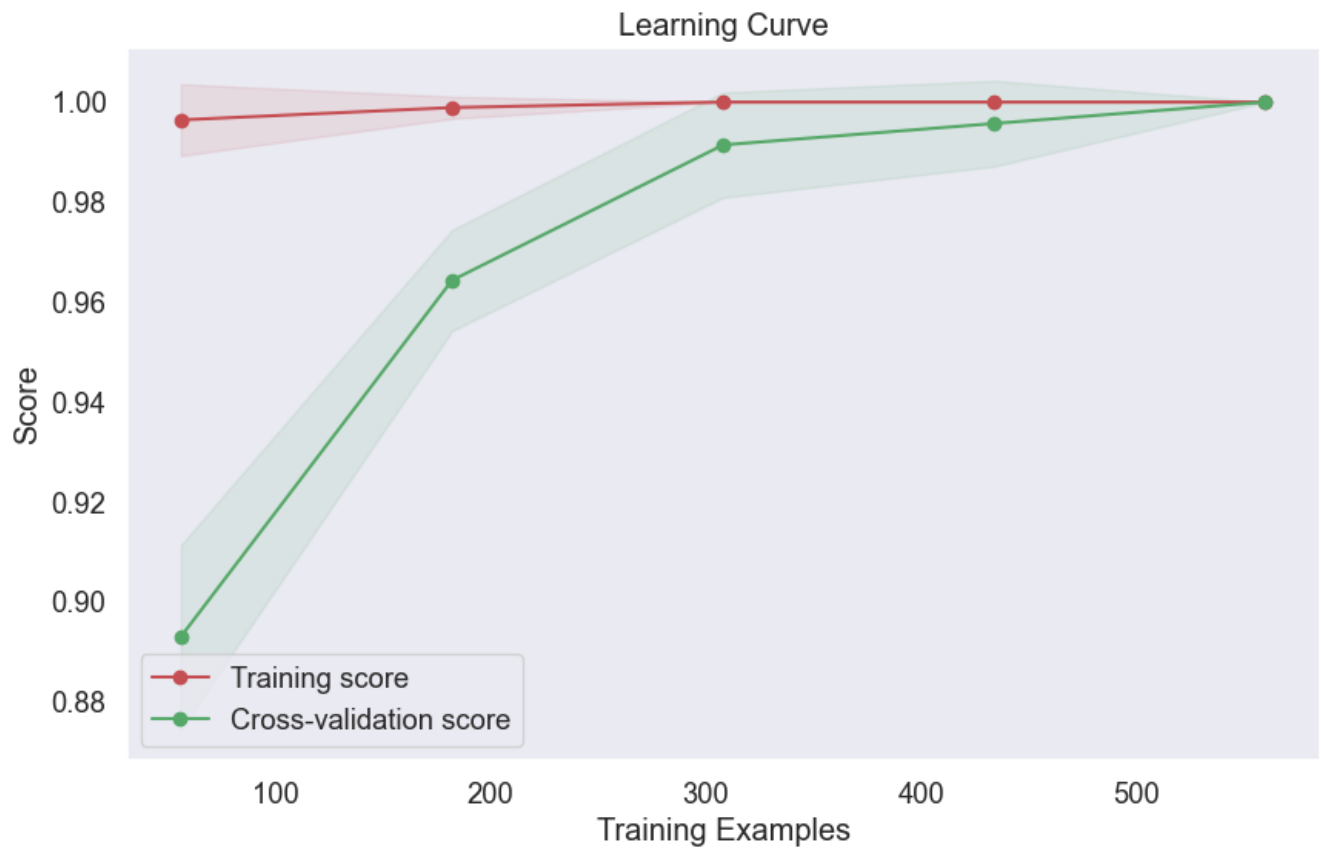

**Figure 10.** Learning Curve of LGBM.

Through our in-depth analysis of the Adaboost model (please refer to Figures 11, 12, and 13), rich information regarding its performance dynamics, and its behavior under changing hyperparameter settings, emerged. Most prominently, a deliberate progression in the learning rates produced incremental improvements in training and validation performance (please refer to Figure 11). What this reveals is the sensititivity of the Adaboost model to changing the learning rate, and the necessity for careful hyperparameter tuning of this hyperparameter in order to maximize model effectiveness. By contrast, fluctuations in minimum child weight showed little impact on model performance, suggesting a comparative insensivity to this specific hyperparameter (please refer to Figure 11).

.

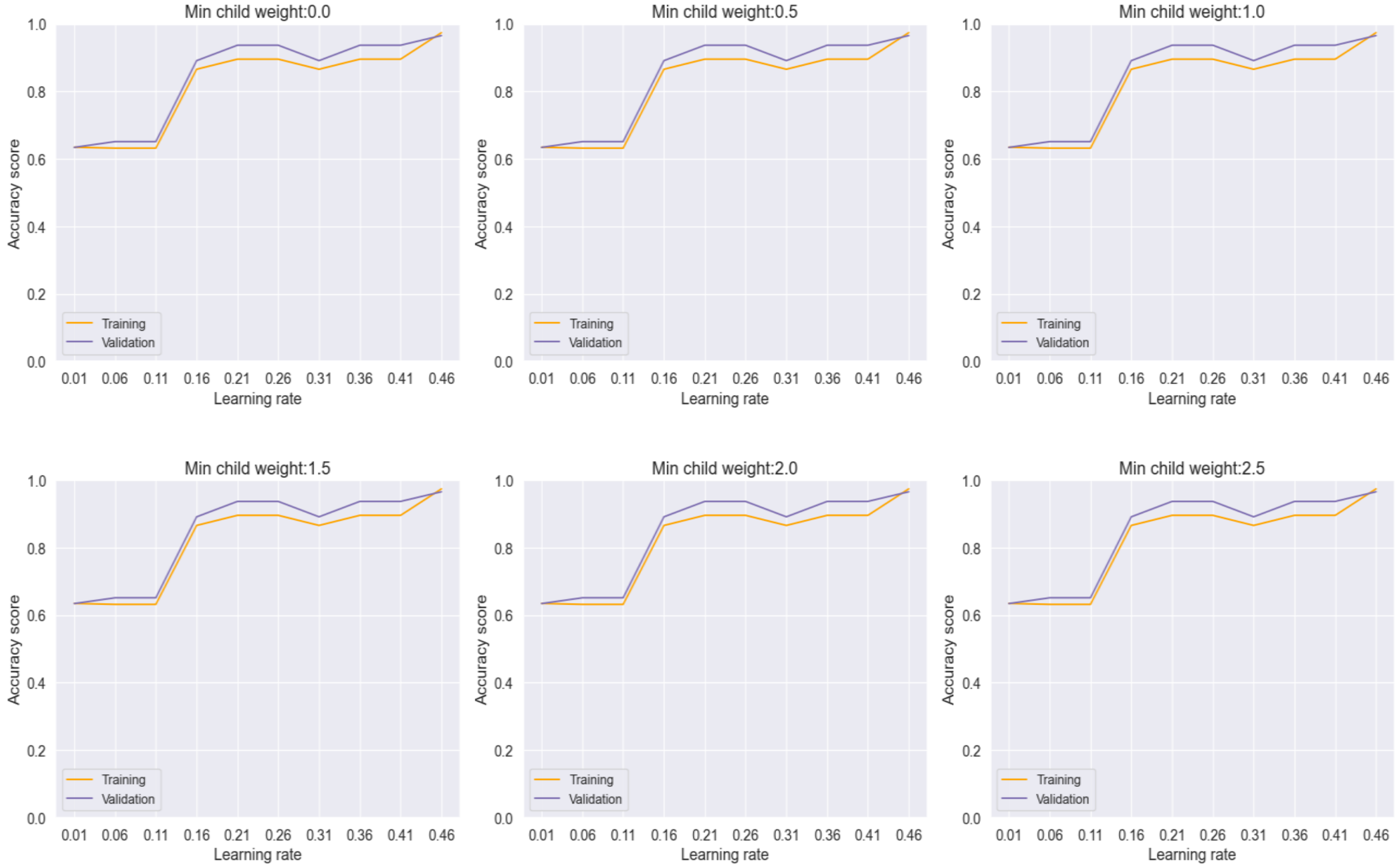

**Figure 11.** Training and Validation plots under consideration of Min child Weight and Learning Rate in AdaBoost.

Although initial concerns about overfitting were present, our thorough analysis of confusion matrices (Figure 12) produced positive indications about model stability. Despite fluctuations in performance statistics, the Adaboost model showed consistent performance in both training and testing phases, thus confirming its dependability and generalizability. In addition, even in cases where model convergence showed decreasing returns with increased datasets (Figure 13), a minor improvement in performance could be seen, with positive implications for model robustness through data augmentation.

The present work clarifies the complex interrelationship between hyperparameters and model performance, providing important insights for both practitioners and researchers. By clarifying model sensitivity to specific parameter settings, our analysis enables wiser decision-making in hyperparameter selection, with a direct improvement in predictive performance. In addition, these observations reiterate model refinement through iterative improvement, supporting the imperative for continuous evaluation and realignment for maximization of model effectiveness.

.

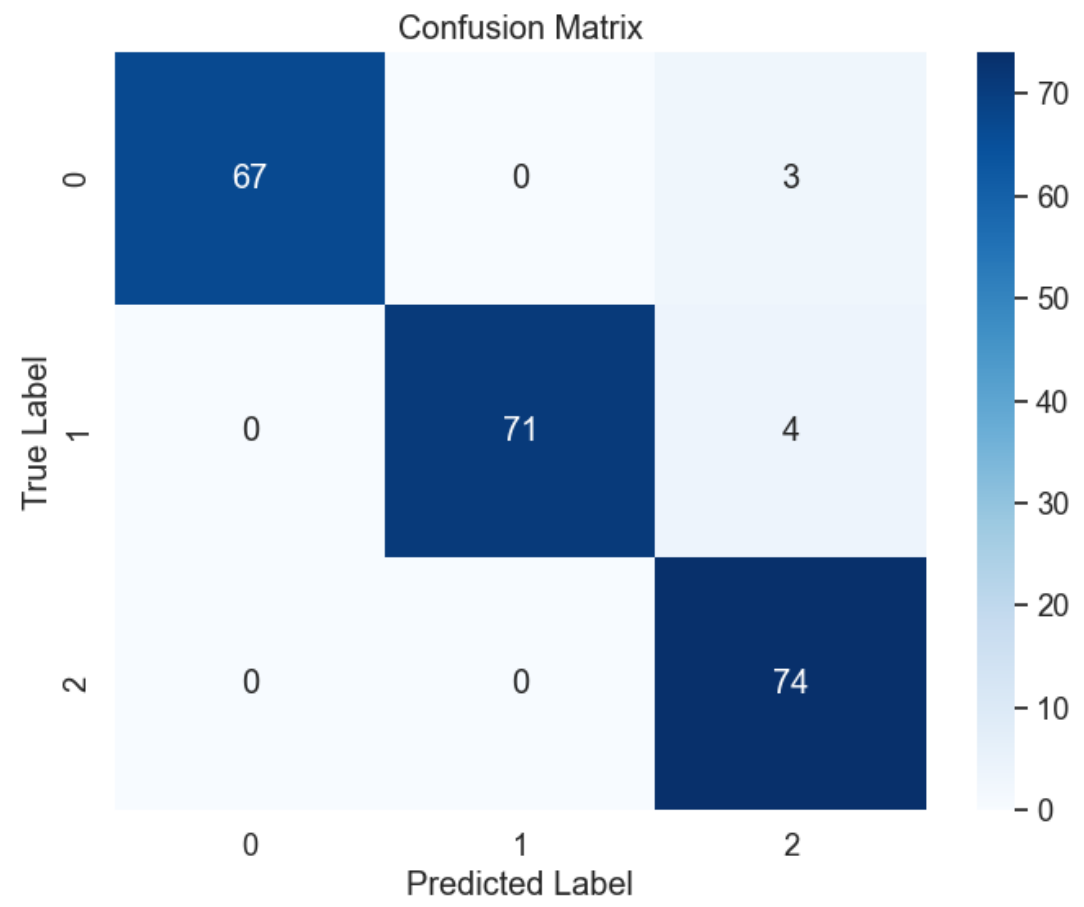

**Figure 12.** Confusion Matrix of AdaBoost.

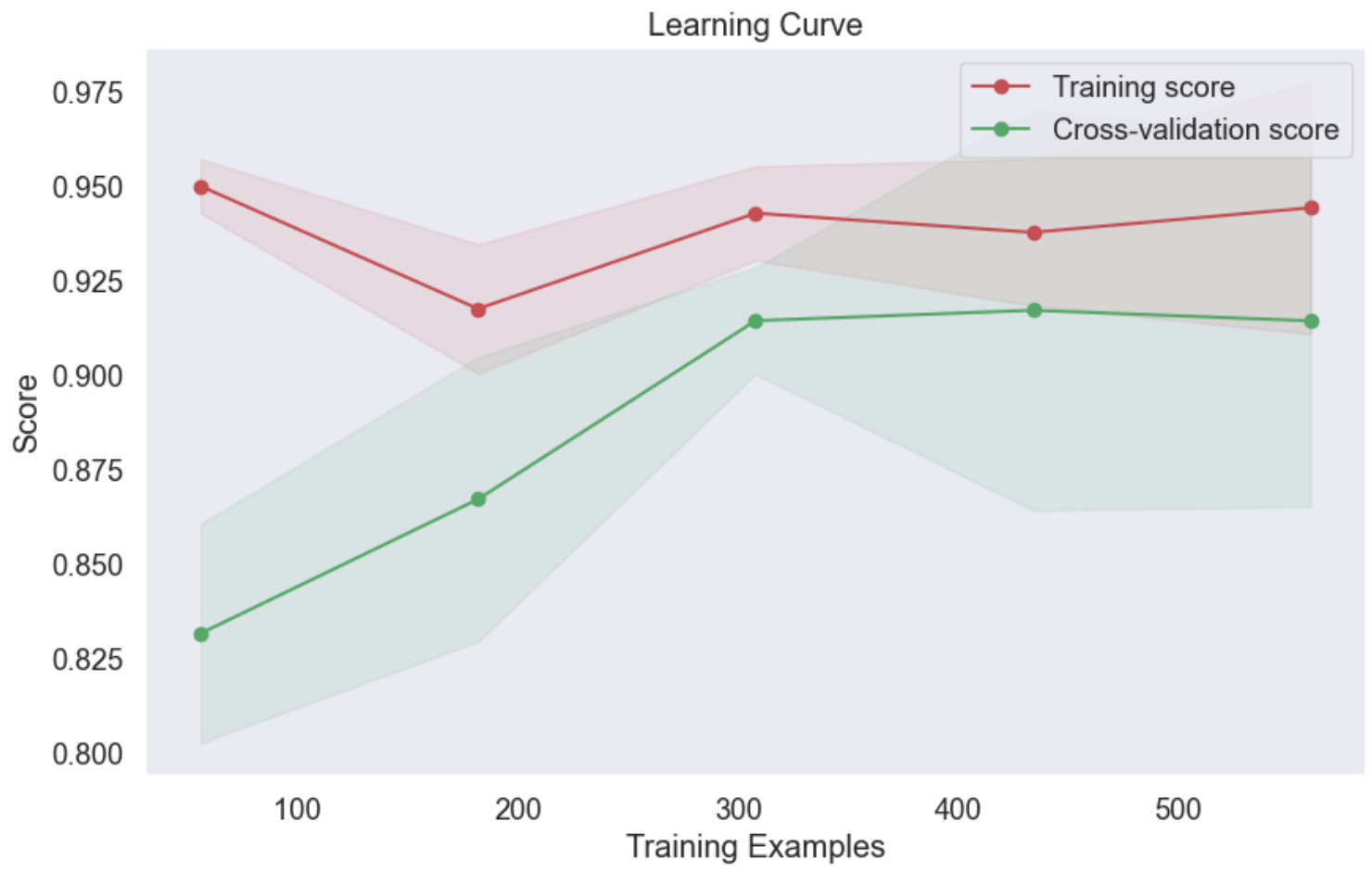

**Figure 13.** Learning Curve of AdaBoost.

A comprehensive analysis of logistic regression model (Figure 14, Figure 15, Figure 16) produced important observations about its performance statistics, confirming its efficacy as a reliable mechanism for predicting lung cancer stages. Most importantly, logistic regression model showed flawless performance in terms of all evaluation factors, underlining its dependability and accuracy in identifying complex trends in the dataset (Figure 14, Figure 15, Figure 16). Notably, concerns about overfitting and underfitting have been carefully addressed, and therefore, allowing for strong generalizability of the model to new datasets.

Moreover, evaluation of the confusion matrix and learning curve supported the model's high performance, with constant values for accuracy, precision, recall, and F-1 score (Figure 14, Figure 15, Figure 16). Overall, these observations confirm the effectiveness of logistic regression model in striking a wise balance between model complexity and generalizability, and in creating a harmonious state that ensures effective prediction in a diverse range of patient groups

.

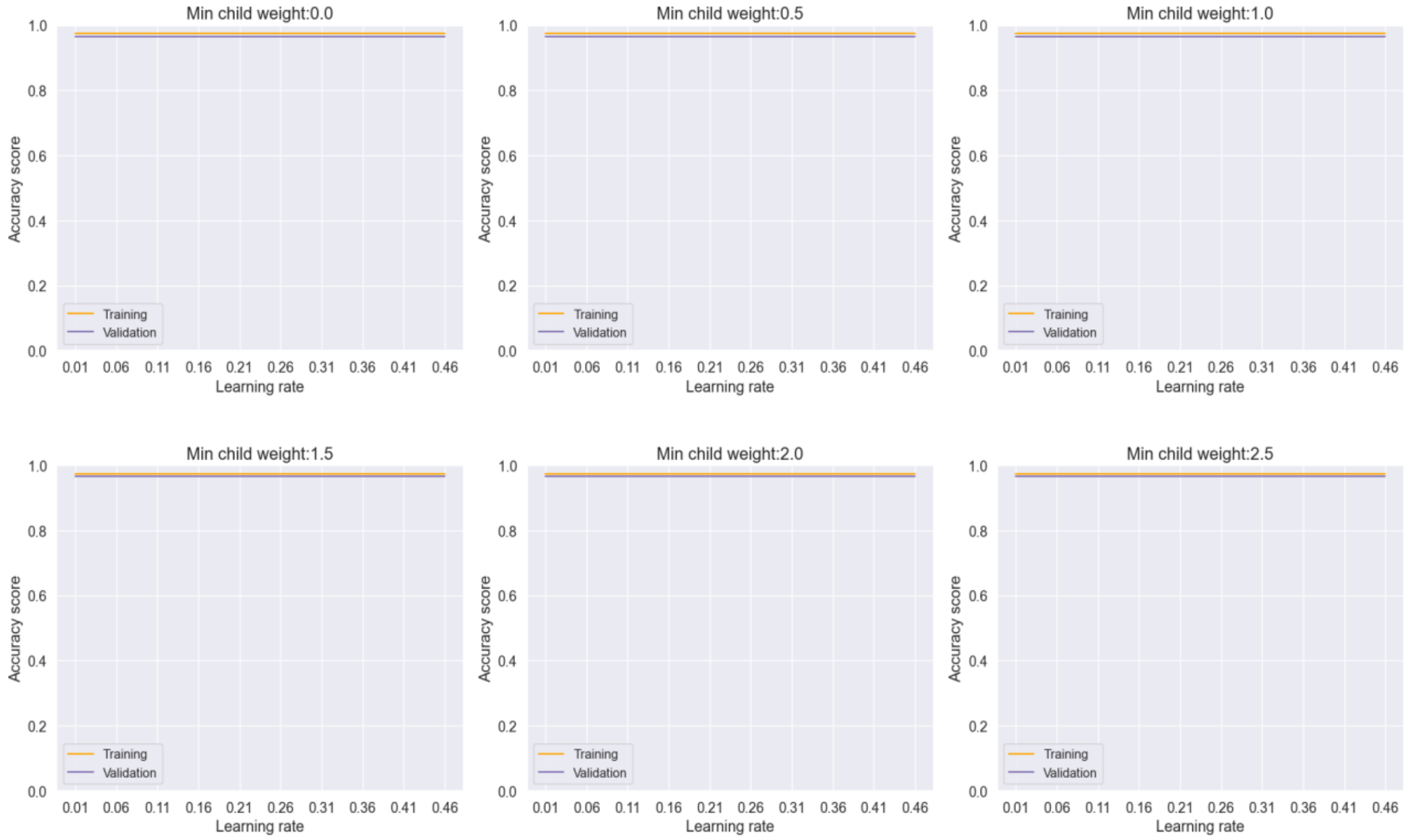

**Figure 14.** Training and Validation plots under consideration of Min child Weight and Learning Rate in Logistic Regression.

Moreover, the flawless accuracy of the logistic regression model underlines its value as a reliable diagnostic tool in clinical settings. By providing accurate and understandable forecasts to medical professionals, such a model enables sound decision-making and aids in personalized therapy for lung cancer patients.

In conclusion, the perfect performance of the logistic regression model, in addition to overfitting and underfitting concerns, underlines its value in predictive modeling in oncologic diagnostics. Its reliability and interpretability make it an inalienable tool in practice, allowing professionals to gain critical information about patient prognosis and therapy planning

.

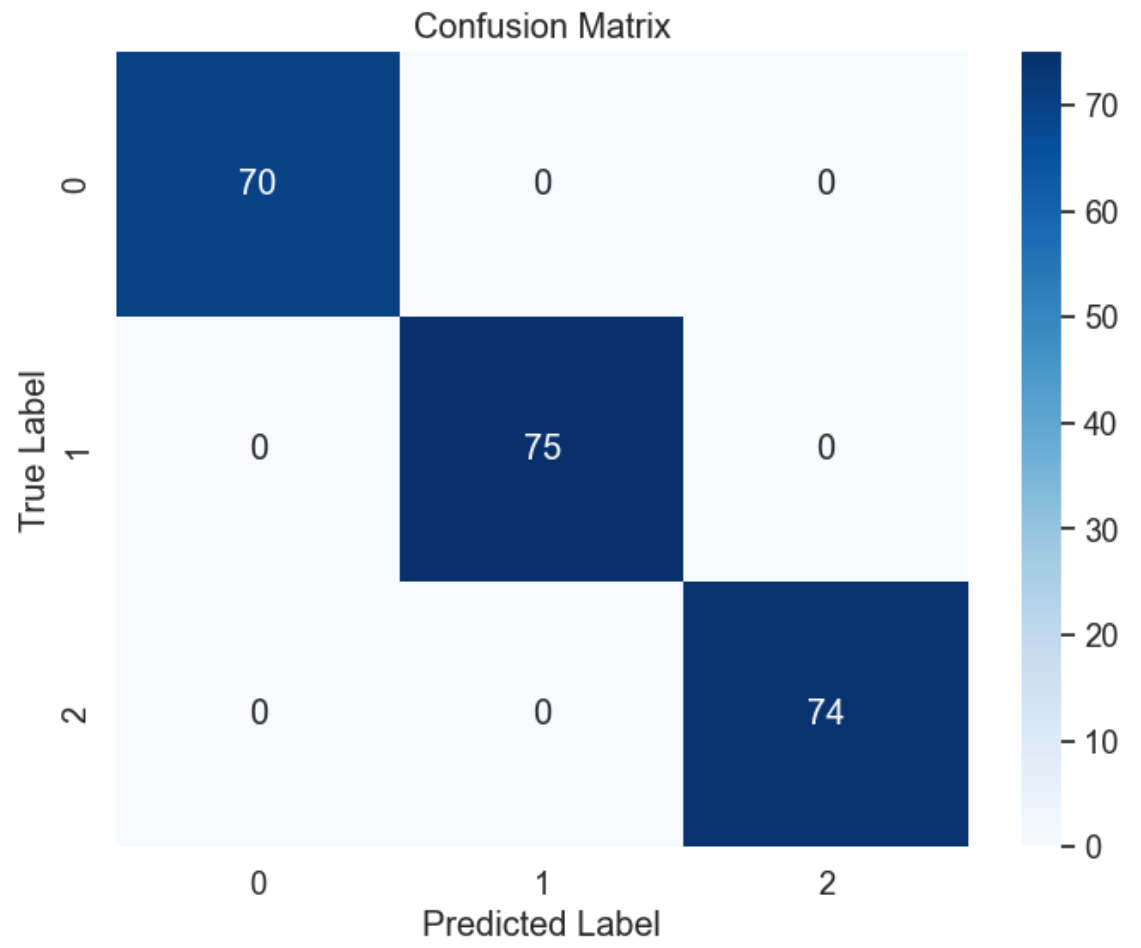

**Figure 15.** Confusion Matrix of Logistic Regression.

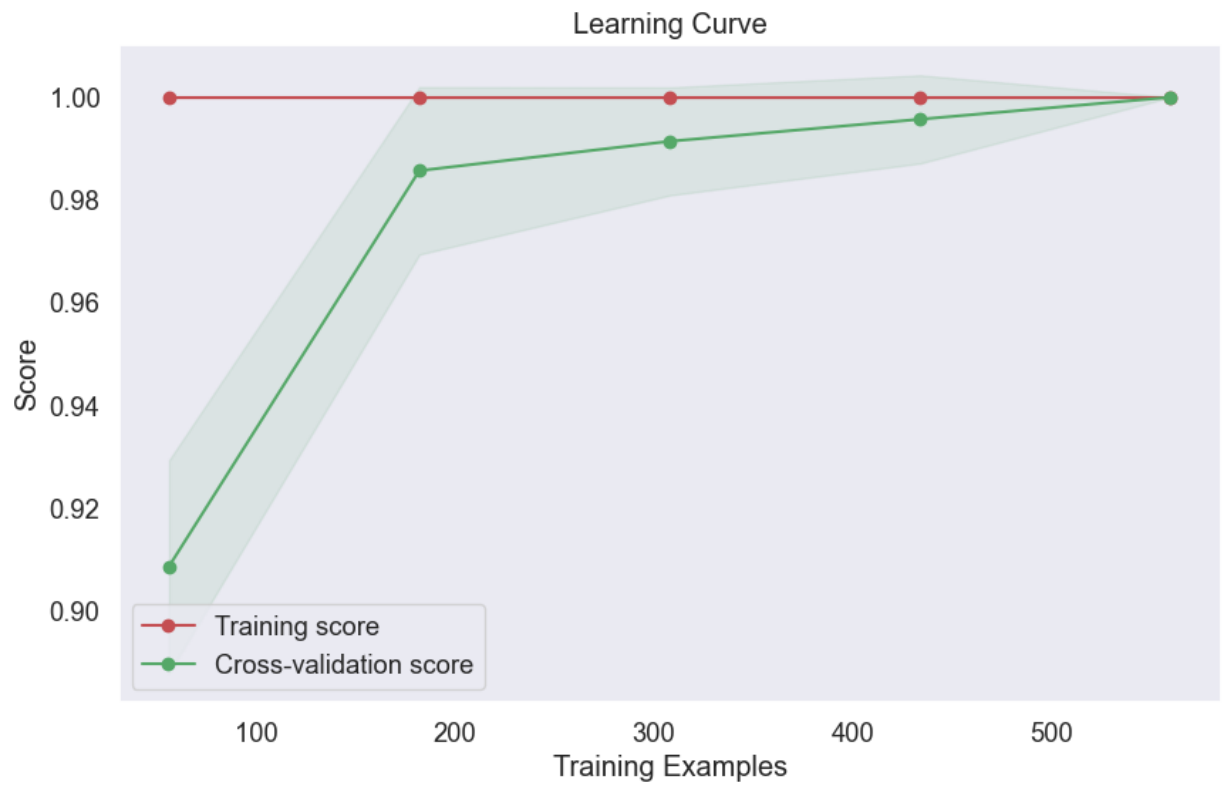

**Figure 16**. Learning Curve of Logistic Regression.

The exploration of the decision tree model (***Figure 17***, ***Figure 18***, ***Figure 19***) provided nuanced insights into its performance dynamics, elucidating its strengths and limitations in lung cancer level classification. Despite variations in learning rates and minimum child weights, the decision tree model exhibited stable performance, achieving an accuracy of 92.69% . This consistency underscores the model's robustness in navigating complex feature spaces and making accurate predictions across diverse patient profiles. However, our analysis revealed that errors primarily occurred in medium and high-level lung cancer classifications, suggesting potential challenges in accurately discerning between these categories. While the decision tree model demonstrated proficiency in distinguishing between lower-level classifications, further refinement may be necessary to improve its performance in higher-level categories.

Interestingly, our findings also highlighted the influence of dataset size on model performance, with optimal performance observed when using fewer datasets (***Figure 19***). This observation suggests that a more focused dataset may facilitate clearer decision boundaries and enhance model interpretability, warranting further investigation into the impact of dataset characteristics on model performance.

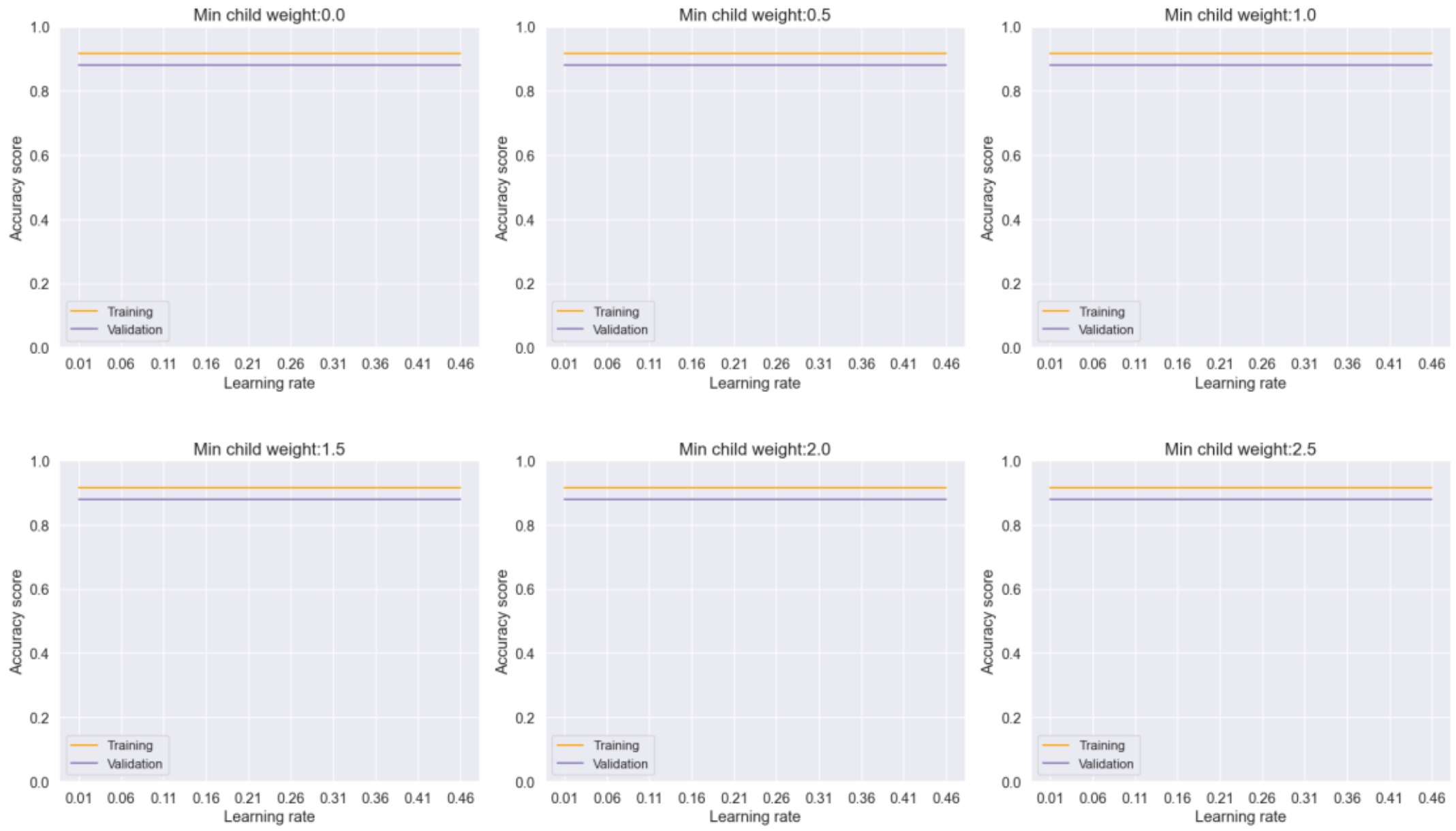

**Figure 17.** Training and Validation plots under consideration of Min child Weight and Learning Rate in Decision Tree.

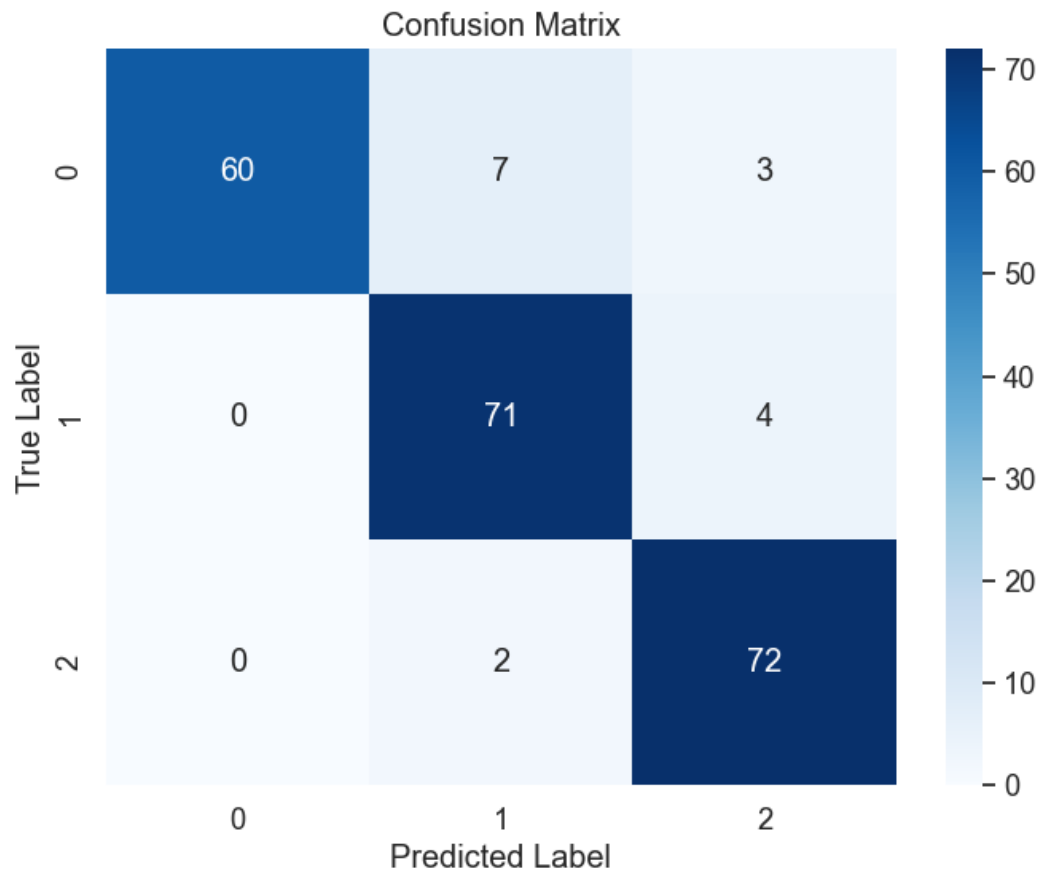

**Figure 18.** Confusion Matrix of Decision Tree.

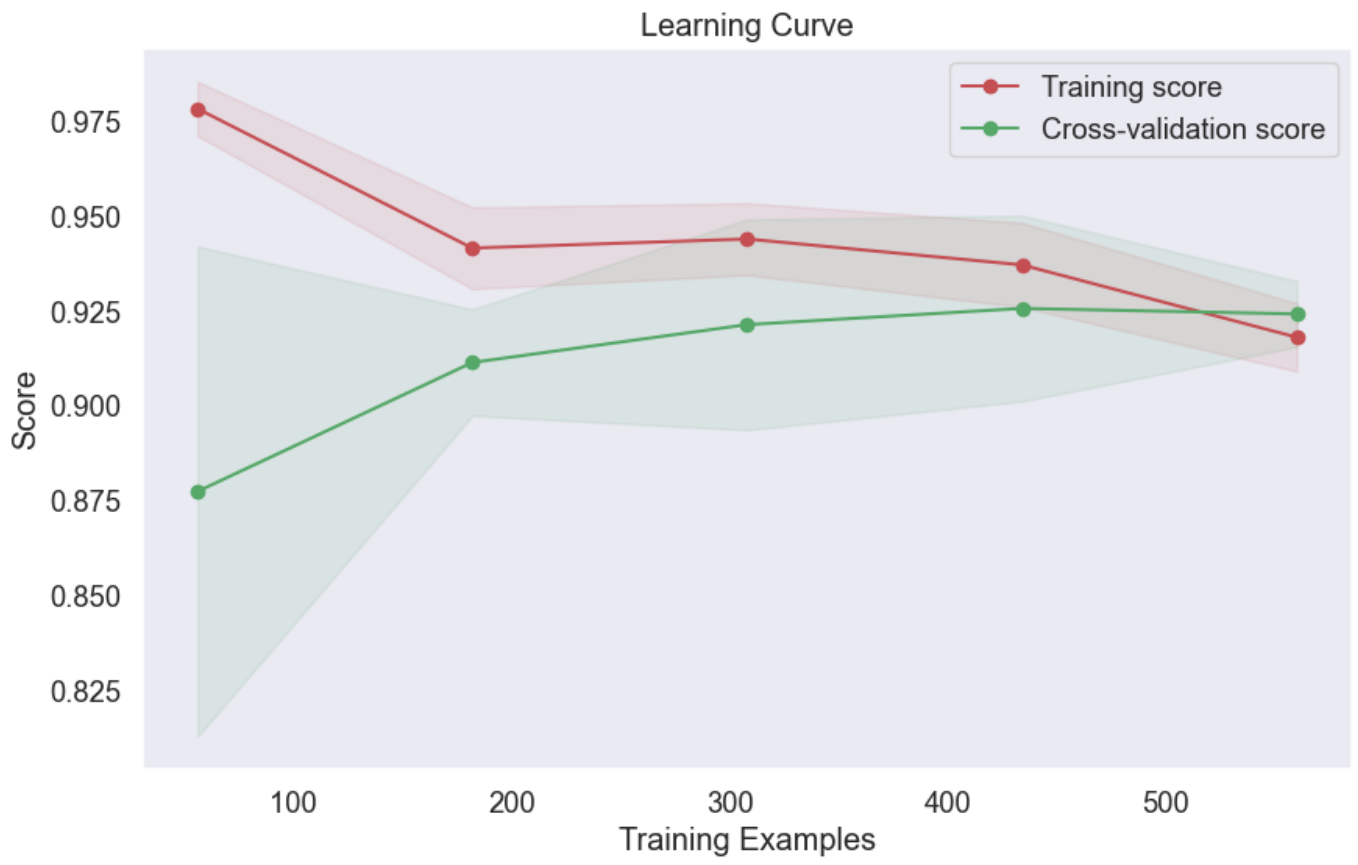

**Figure 19.** Learning Curve of Decision Tree.

The comprehensive evaluation of the Random Forest model (Figure 20, Figure 21, Figure 22) showed high effectiveness, with high accuracy and no significant overfitting concerns. Our thorough investigation, including a review of confusion matrices and learning curves, confirmed the model's solidity and its generalizability to a variety of datasets.

Our investigation of the CatBoost and k-NN models (Figure 23, Figure 24, Figure 25, Figure 26, Figure 27, Figure 28) showed similar performance in terms of accuracy and dependability, similar to that seen with the Random Forest model. All models showed uniform accuracy and dependability, with high efficacy in predicting lung cancer stages. Notably, our investigation showed increased performance with increased datasets, and therefore, with room for future investigation and improvement. The flawless performance of the Random Forest, CatBoost, and k-NN models attests to the effectiveness of ensemble approaches and neighbor techniques in dealing with complex classification problems. By leveraging the synergistic capabilities of several decision trees or neighbor examples, these models effectively reveal complex structures in the data, and therefore, make accurate and reliable forecasts

.

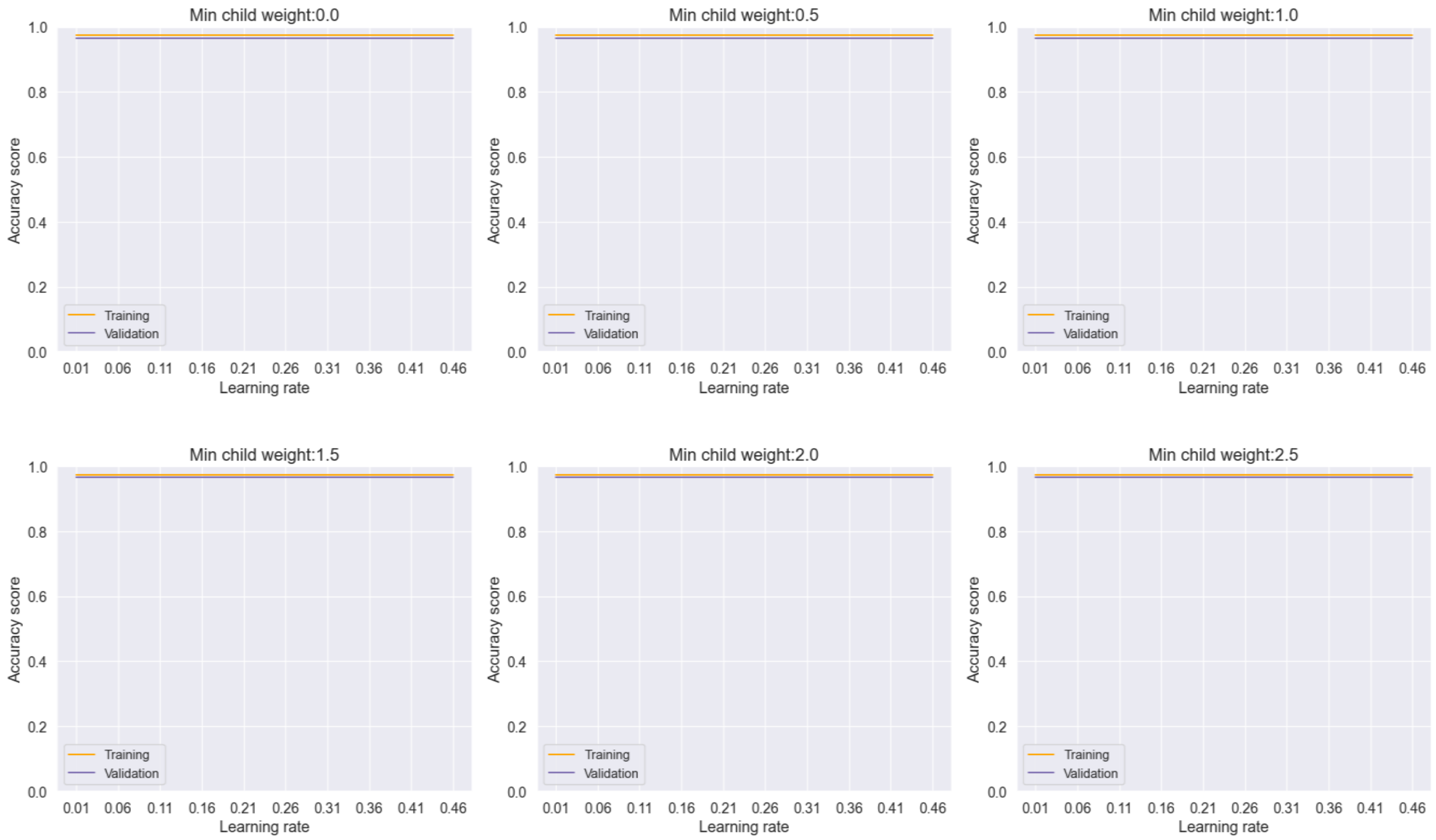

**Figure 20.** Training and Validation plots under consideration of Min child Weight and Learning Rate in Random Forest.

Moreover, the insensibility of such models towards overfitting concerns underlines the effectiveness of ensemble approaches in resolving model complexity-related challenges. In medical settings, such insensibility is specifically significant, with high predictive accuracy being critical for guiding interventions in patient care.

Our work, therefore, re-affirms CatBoost, Random Forest, and k-NN's performance over the classification phases of lung cancer. By investigating their strengths and weaknesses, our work helps in ongoing development of predictive modeling approaches in oncologic diagnostics for enhancing patient care and enhancing our understanding of such a complex disease.

.

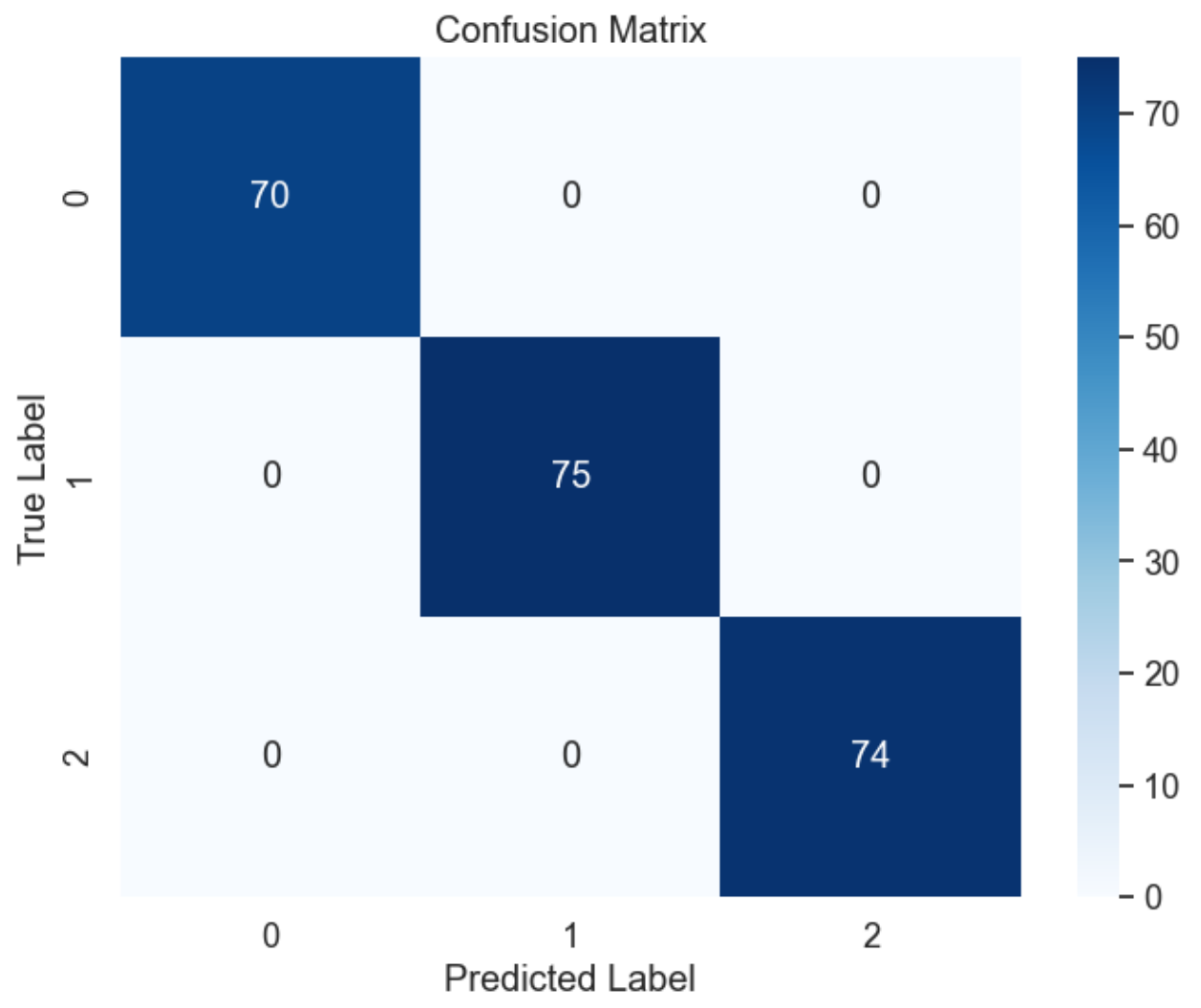

**Figure 21.** Confusion Matrix of Random Forest.

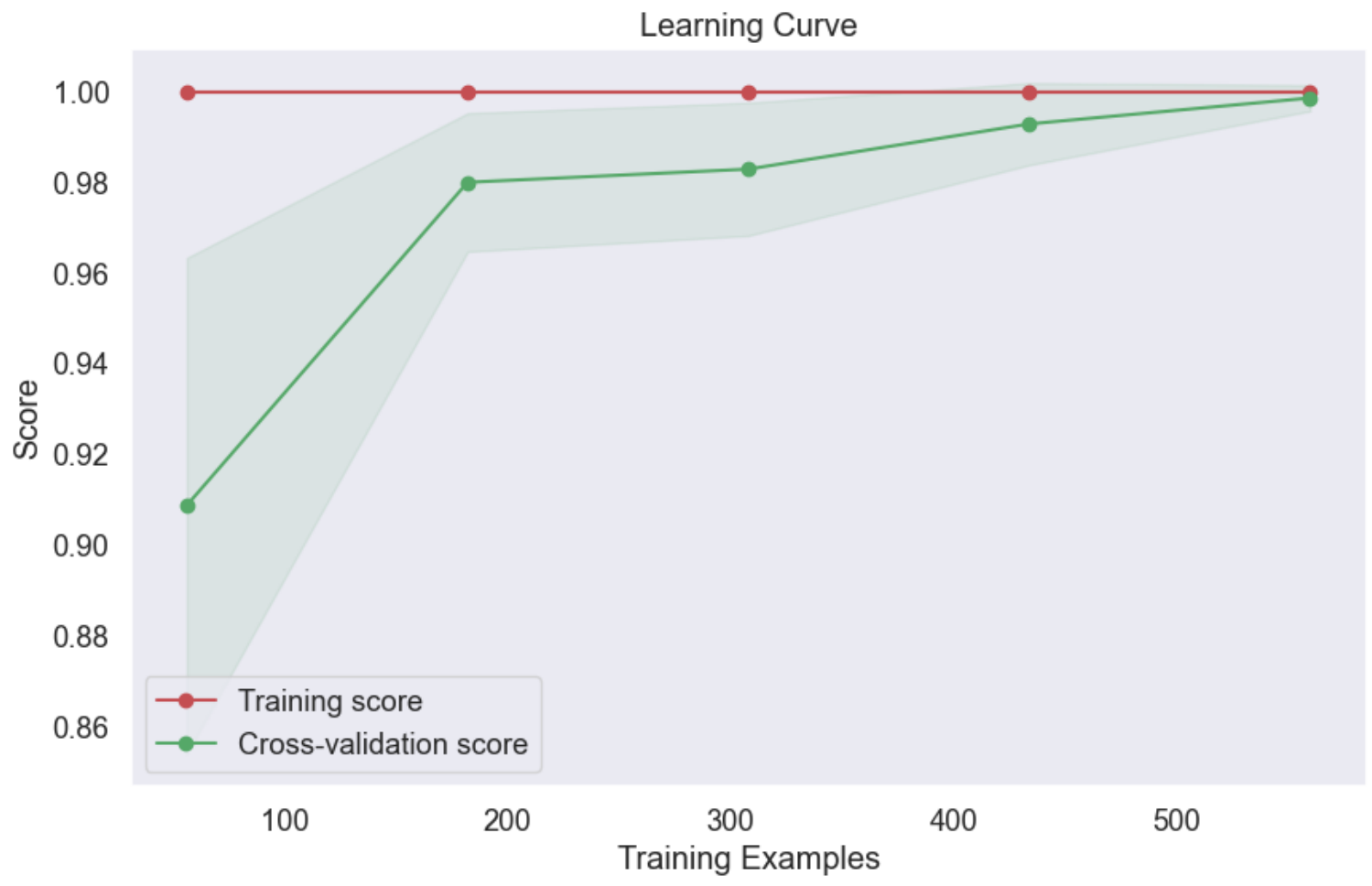

**Figure 22.** Learning Curve of Random Forest.

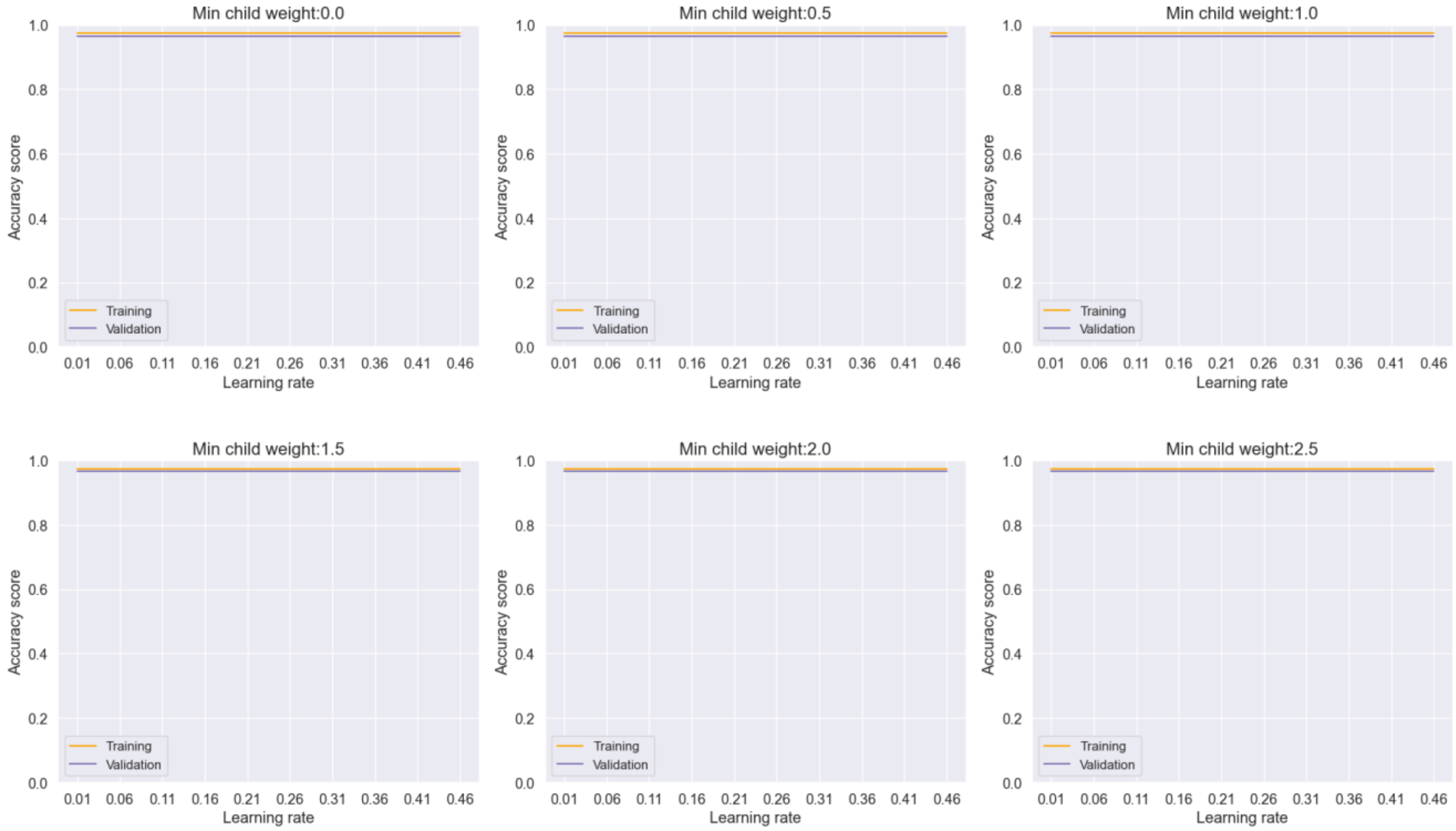

**Figure 23.** Training and Validation plots under consideration of Min child Weight and Learning Rate in CatBoost.

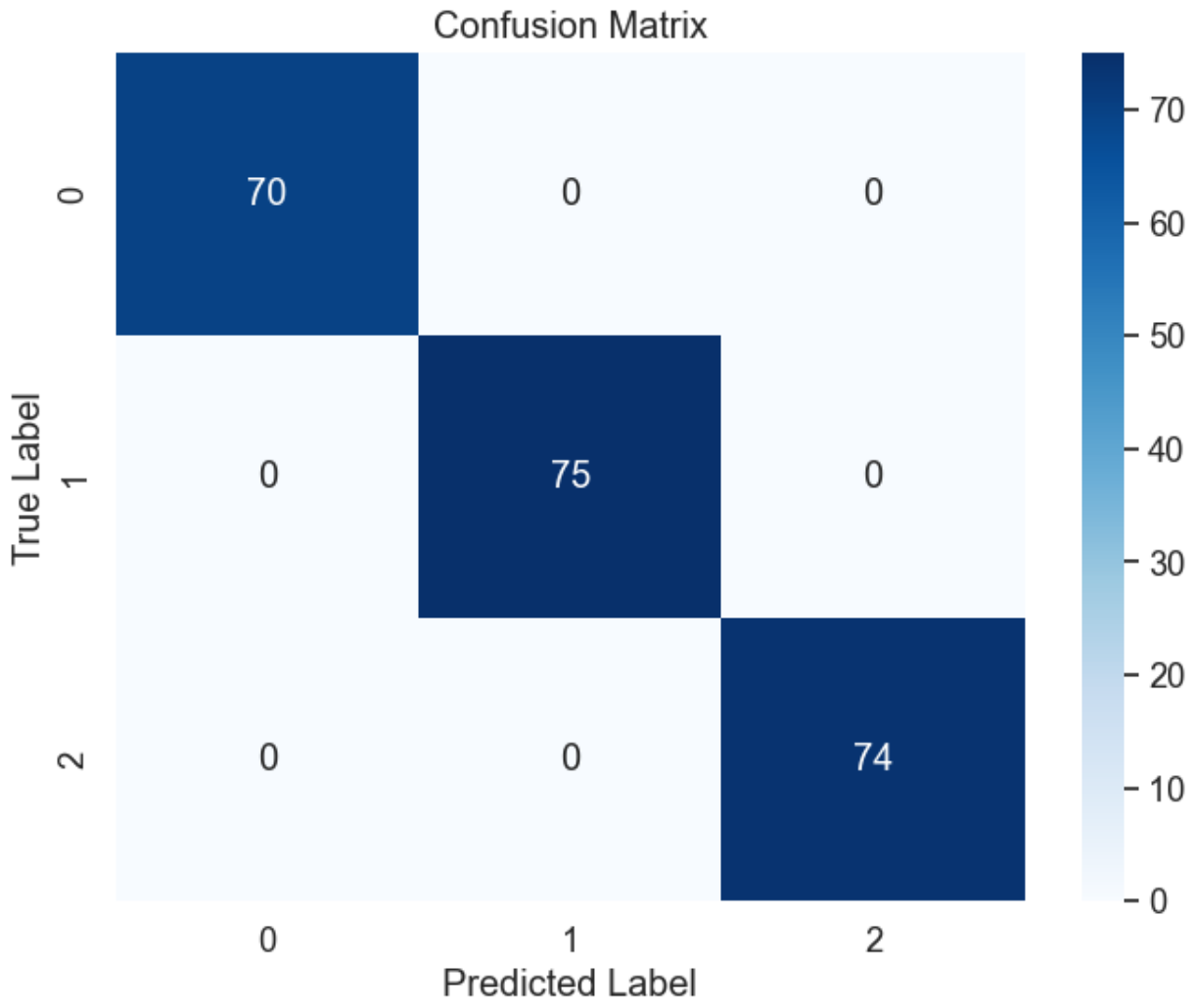

**Figure 24.** Confusion Matrix of CatBoost.

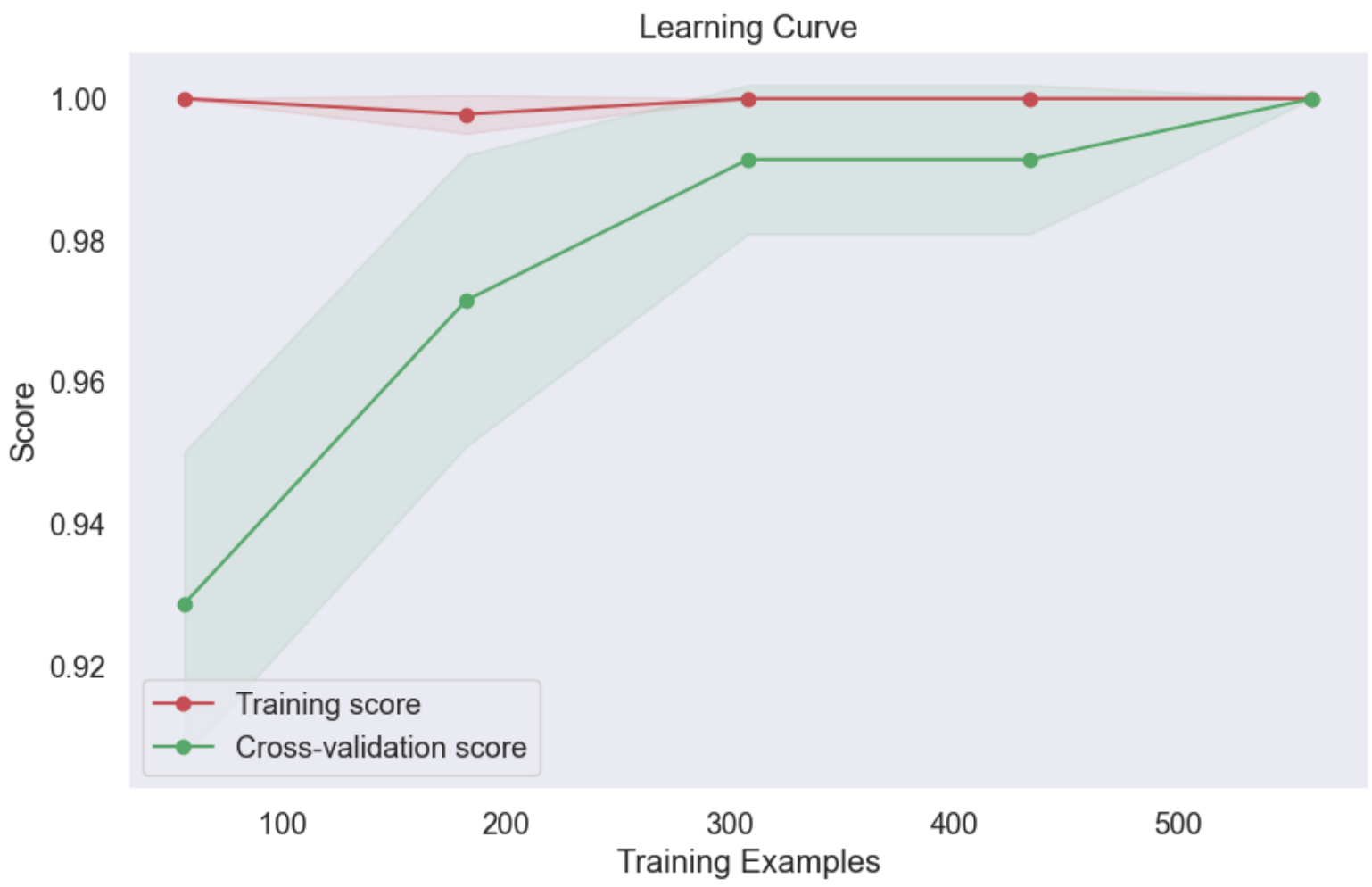

**Figure 25.** Learning Curve of CatBoost.

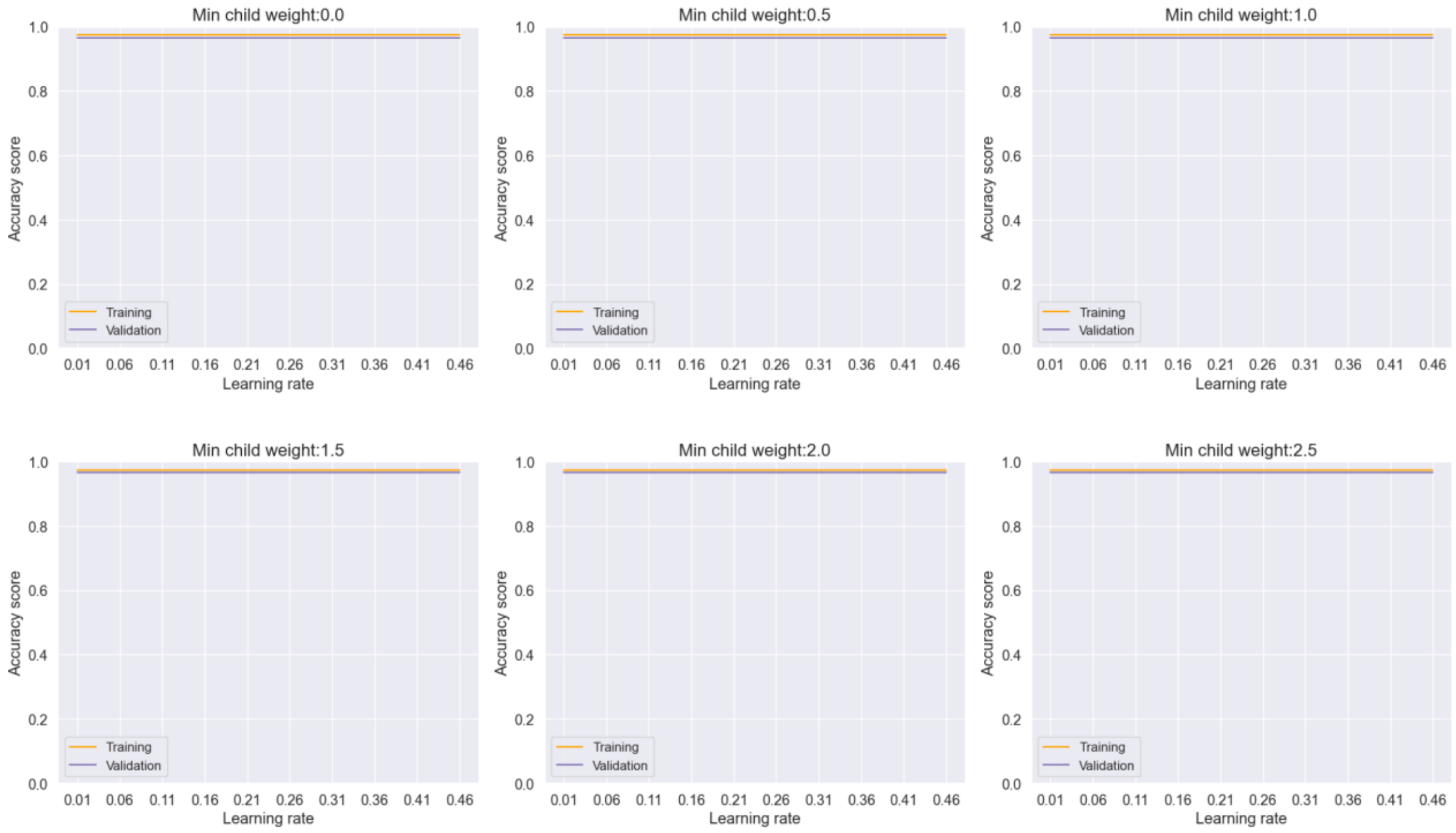

**Figure 26.** Training and Validation plots under consideration of Min child Weight and Learning Rate in k-NN.

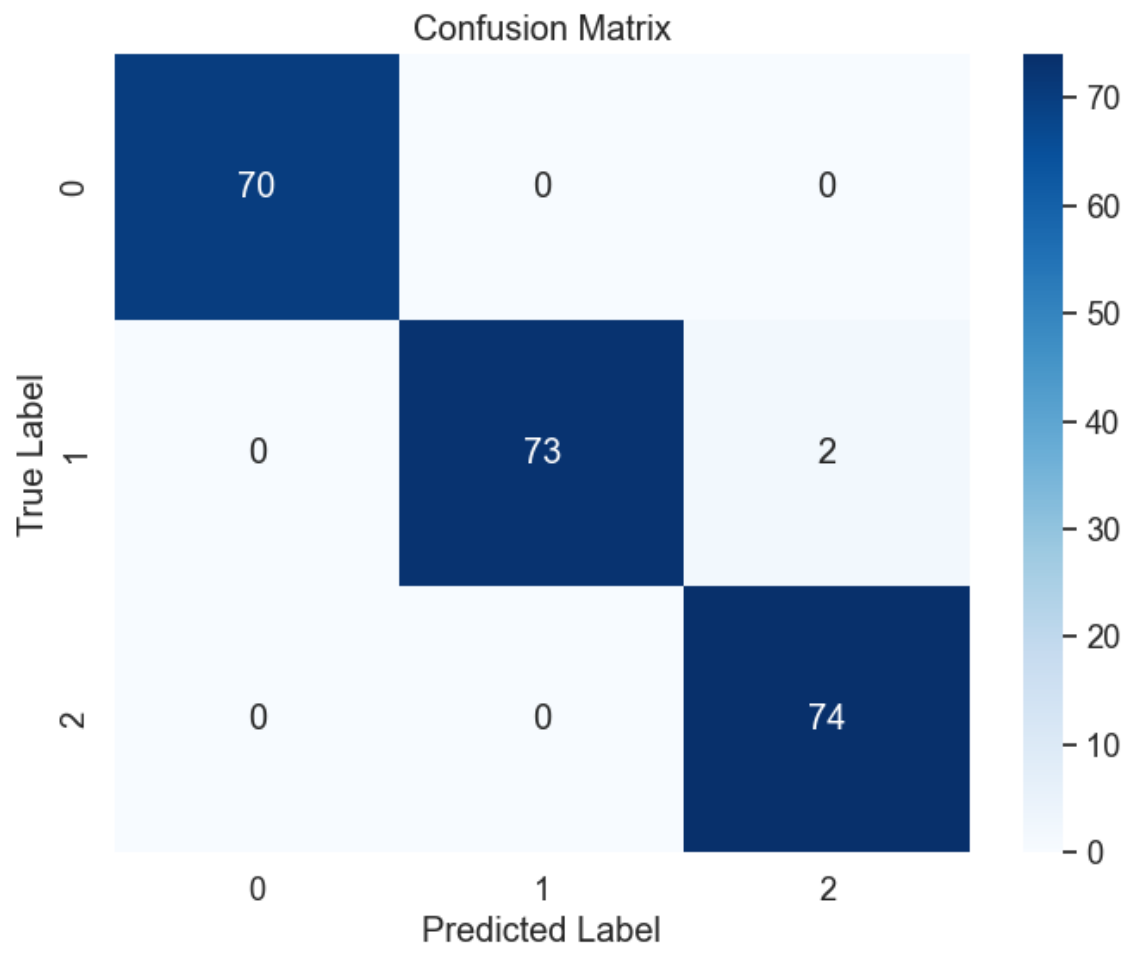

**Figure 27.** Confusion Matrix of k-NN.

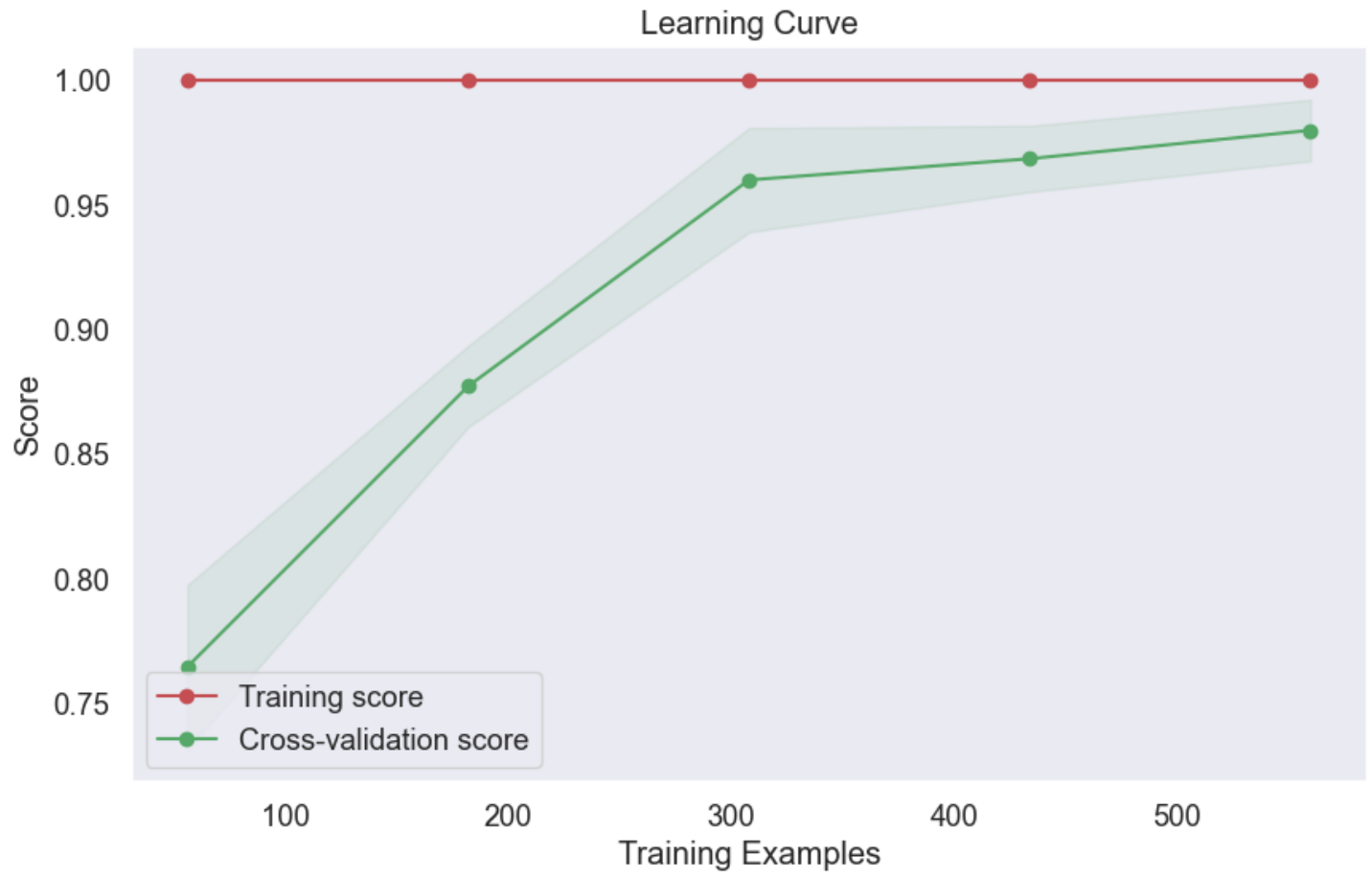

**Figure 28.** Learning Curve in k-NN.

The integration of Deep Neural Networks (DNN) (see Figure 29 and Figure 30) in this study was designed to leverage their inbuilt capability for explaining complex relationships between targets and features. Despite their potential for discovering complex patterns in the dataset, our observations showed that conventional machine learning (ML) models outperformed DNN in lung cancer level classification. Despite its consistent performance during both training and validation phases, with perfect classification of lung cancer level, DNN failed to outdo conventional ML models in efficacy. This observation reflects the nuanced dynamics between model architectures and inbuilt complications in the dataset.

A variety of factors could contribute to performance gaps in our investigation. First, relatively small datasets could limit DNNs' ability to maximize feature extraction and representation capabilities. In addition, computational requirements for training DNNs necessitate larger datasets and long training times for optimized performance, factors not addressed in our studies' scope. In spite of such challenges, use of DNNs reflects continued efforts towards developing sophisticated modeling techniques in oncologic diagnostics. Future studies can gain value through use of larger datasets and hyperparameter search for optimized performance in leveraging capabilities of DNNs in lung cancer level classification..

Future research endeavors may benefit from larger datasets and more extensive hyperparameter tuning to fully unlock the potential of DNNs in lung cancer level classification.

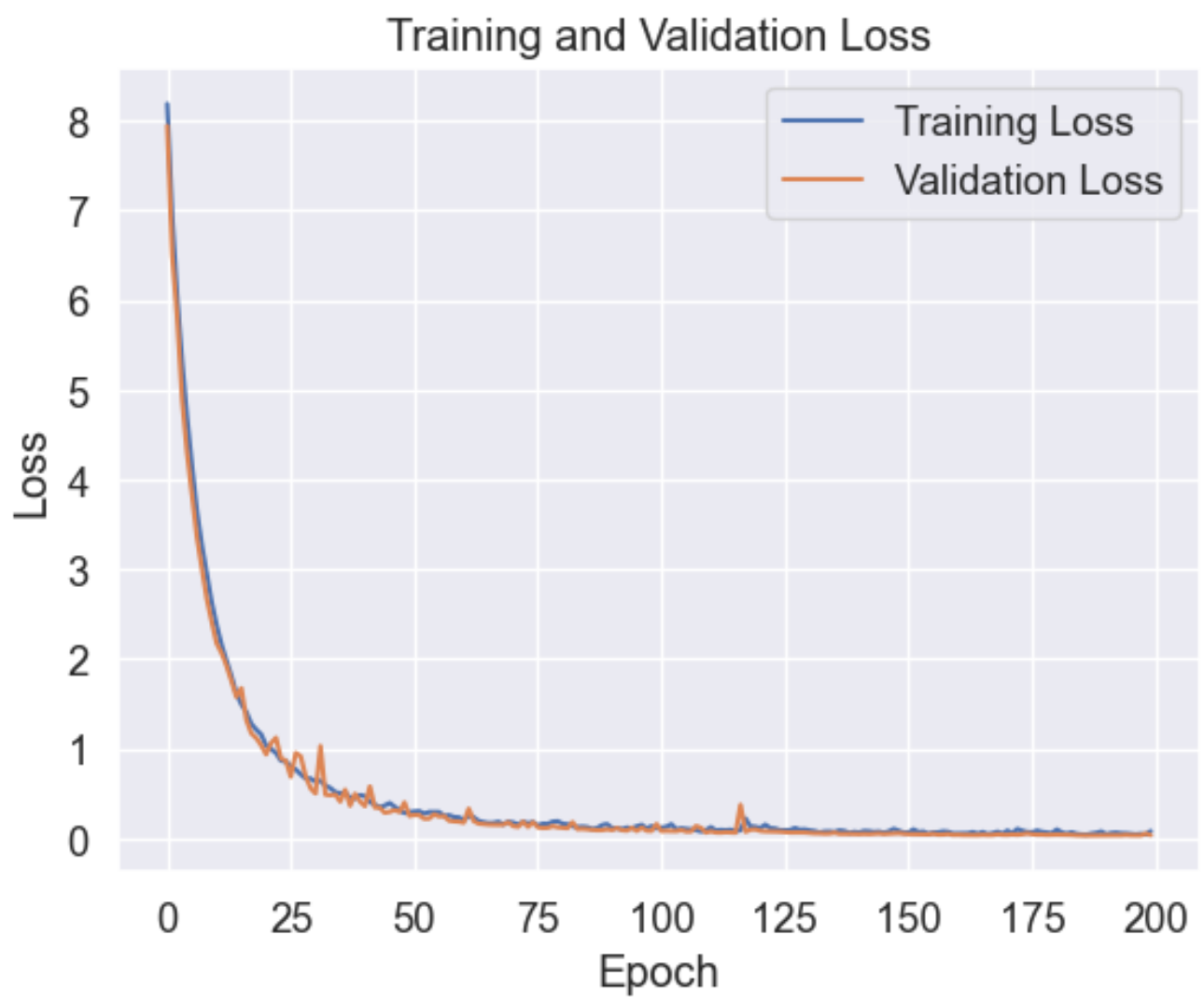

**Figure 29.** Training vs. Validation of DNN.

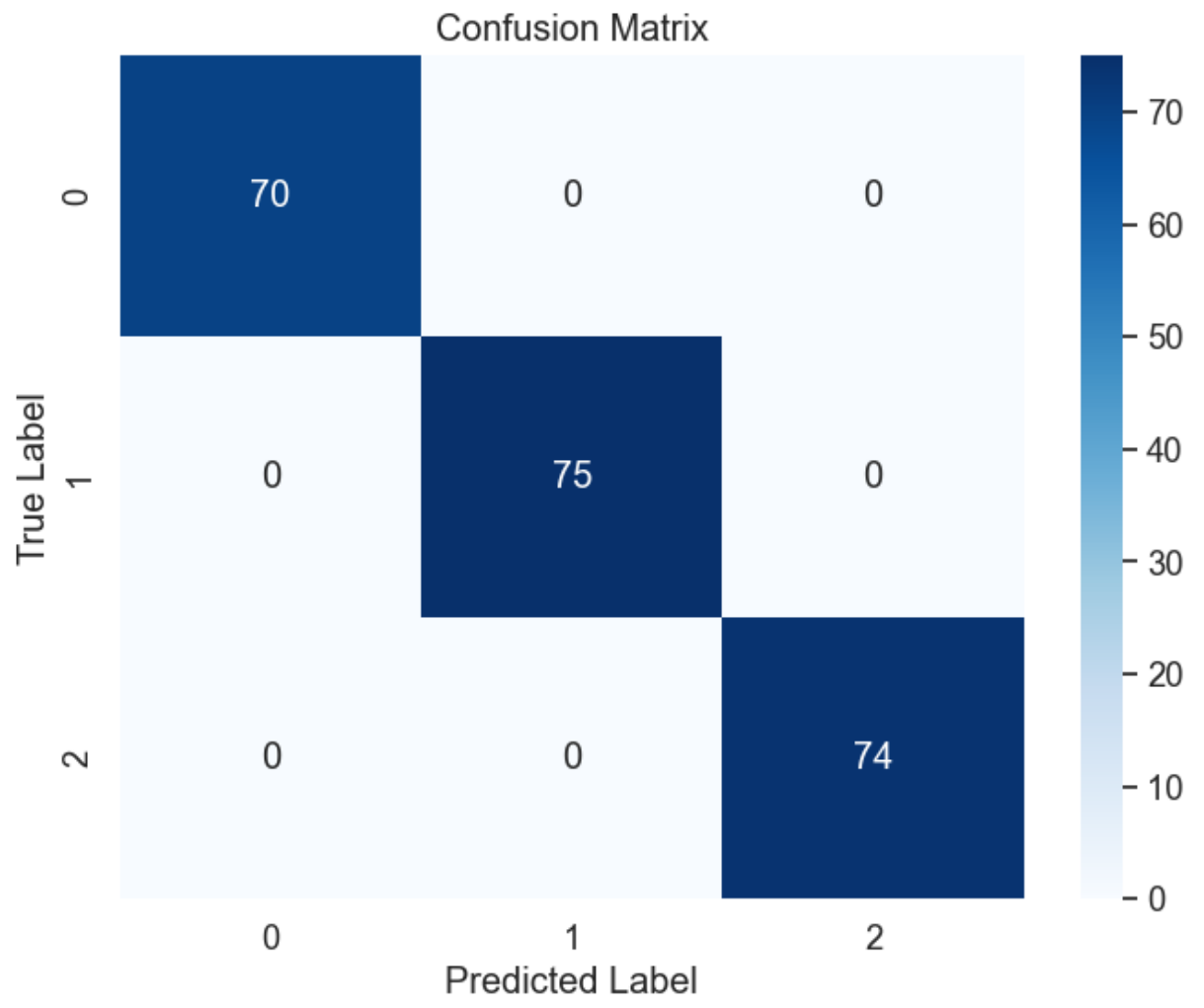

**Figure 30.** Confusion Matrix of DNN.

The comprehensive comparative analysis represented in Figure 31 yields significant information regarding the performance of various machine learning algorithms in lung cancer stage classification. By careful evaluation of key performance statistics, such as accuracy, precision, recall, and F-1 score, represented in Figure 32, our analysis reveals that DNN, XGBoost, LGBM, Logistic Regression, CatBoost, and Random Forest performed best.

The comparative analysis serves as a critical benchmark for clinicians and researchers, providing them with a basis for sound decision-making in model selection for lung cancer prediction-related tasks. By combining performance statistics for a range of disparate models, we present a critical evaluation of individual model strengths and weaknesses, and in consequence, enable determination of best-fit modeling approaches for specific clinical scenarios

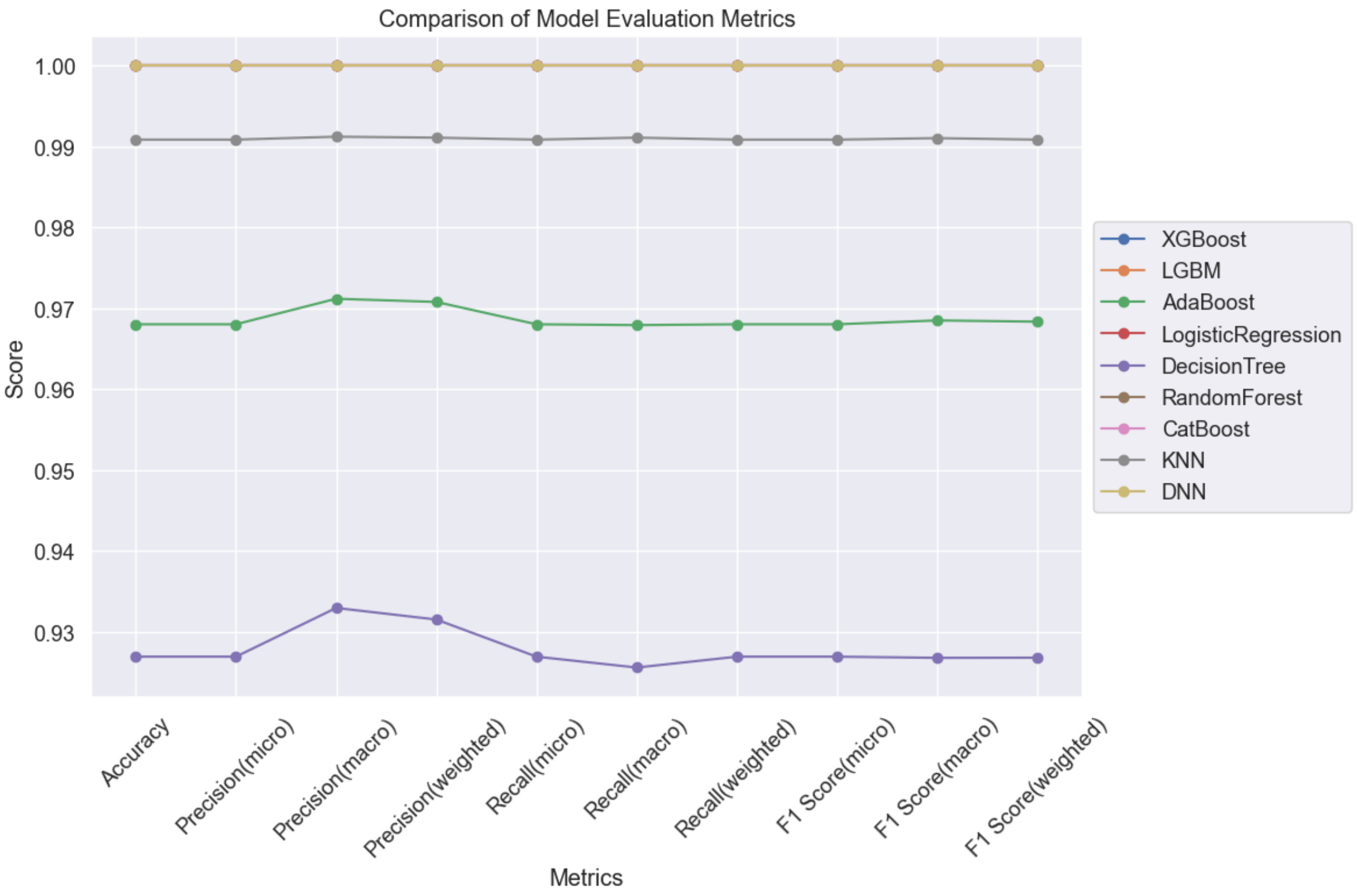

**Figure 31.** Comparison of ML Models for Lung Cancer Prediction.

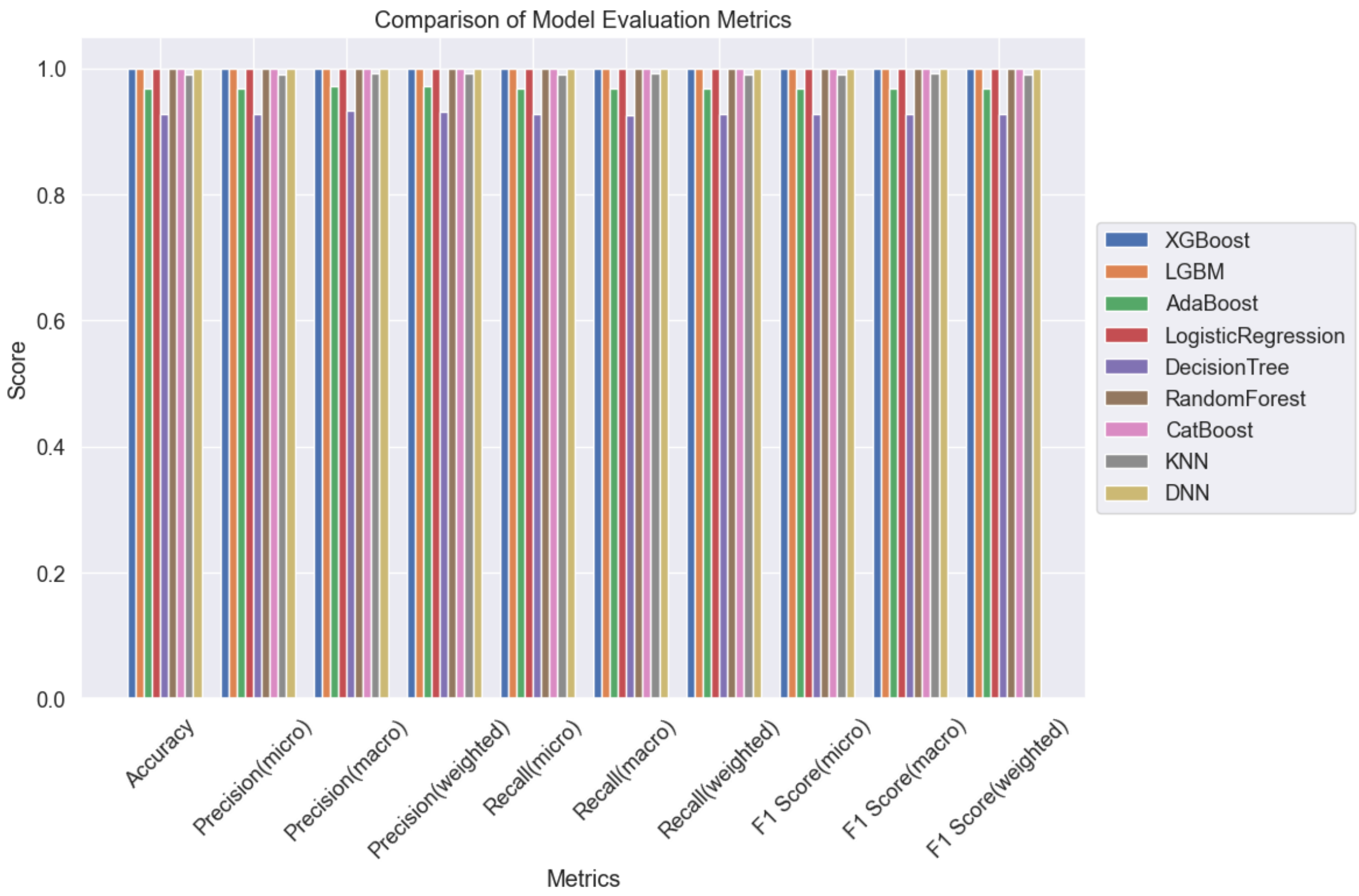

**Figure 32.** Comparison of 9 ML Models for Lung Cancer Prediction.

Furthermore, the acknowledgement of high-performance models accentuates the diversity of predictive modeling approaches utilized in oncological diagnostics. Certain models, for instance, such as Deep Neural Networks (DNN), excel at discovering complex patterns in the data, whereas others, such as logistic regression and ensemble approaches, prioritize providing strong and unambiguous predictions. This diversity accentuates the imperative for investigating a range of modeling approaches in an attempt to gain improvements in predictive accuracy and in terms of clinical utility.

In conclusion, our comparative analysis accentuates the efficacy of such models as DNN, XGBoost, LGBM, Logistic Regression, CatBoost, and Random Forest in predicting lung cancer severity. By a thorough analysis of their performance characteristics, our work promotes ongoing development in predictive modeling approaches in oncological diagnostics, with a view towards enhancing patient care and deepening our shared knowledge of such a complex disease.

## Conclusion

In conclusion, the in-depth analysis of machine learning (ML) models for lung cancer level classification re-emphasizes the critical role played in careful hyperparameter supervision, with specific regard for factors such as minimum child weight and learning rate. By careful analysis of learning curve and confusion matrix, we observed careful consideration for overfitting in all models.

XGBoost and LGBM models performed exemplary, with perfect accuracy, precision, recall, and F-1 score. Notably, careful hyperparameter tweaks, particularly in terms of an increased learning rate over 0.06, helped in performance improvement with no overfitting.

Adaboost performed with increased improvements in performance with increased learning rates, but minimum child weight changes showed little impact in model performance. Nonetheless, its consistent performance in both training and testing phases reiterated its dependability.

The logistic regression and decision tree model performed consistently, with careful consideration for overfitting and underfitting. Notably, the decision tree model performed best when trained with smaller datasets, and therefore, re-emphasizes the role played in training with datasets.

Random Forest, CatBoost, and k-NN performed perfectly with no overfitting concerns, and therefore, reiterate robust performance in lung cancer level classification.

The incorporation of additional datasets showed potential for additional performance improvement, particularly in k-NN models.

Not withstanding an investigation of Deep Neural Networks (DNN) for feature-target complexity, conventional ML models outperformed deep learning counterparts consistently, with perfect classification of lung cancer level.

In comparative analysis, we identified top performers, including DNN, XGBoost, LGBM, Logistic Regression, CatBoost, and Random Forest, in a range of evaluation metrics.

In summary, in this work, the effectiveness of machine learning algorithms in predicting lung cancer severity with high accuracy is clarified. By careful analysis of model parameters and careful evaluation of performance metrics, we achieved high-performance results with reduced overfitting risk. Traditional machine learning algorithms, such as XGBoost and Logistic Regression, ranked high in performance in our evaluated models and showed significant potential for lung cancer prediction in a clinic environment. Not only does this work deepen our understanding of the role of machine learning in oncologic diagnostics, but it opens doors for enhanced patient care through increased accuracy in prognosis and therapeutic planning.

## Reference

[1] E. Dritsas, M. Trigka, Lung Cancer Risk Prediction with Machine Learning Models, Big Data and Cognitive Computing 2022, Vol. 6, Page 139 6 (2022) 139. https://doi.org/10.3390/BDCC6040139.

[2] M.D. Podolsky, A.A. Barchuk, V.I. Kuznetcov, N.F. Gusarova, V.S. Gaidukov, S.A. Tarakanov, Evaluation of Machine Learning Algorithm Utilization for Lung Cancer Classification Based on Gene Expression Levels, Asian Pacific Journal of Cancer Prevention 17 (2016) 835–838. https://doi.org/10.7314/APJCP.2016.17.2.835.

[3] I. El Naqa, Machine learning methods for predicting tumor response in lung cancer, Wiley Interdiscip Rev Data Min Knowl Discov 2 (2012) 173–181. https://doi.org/10.1002/WIDM.1047.

[4] M. Hilario, A. Kalousis, M. Müller, C. Pellegrini, Machine learning approaches to lung cancer prediction from mass spectra, Proteomics 3 (2003) 1716–1719. https://doi.org/10.1002/PMIC.200300523.

[5] S.A. Ajagbe, O.A. Oki, M.A. Oladipupo, A. Nwanakwaugwum, Investigating the Efficiency of Deep Learning Models in Bioinspired Object Detection, International Conference on Electrical, Computer, and Energy Technologies, ICECET 2022 (2022). https://doi.org/10.1109/ICECET55527.2022.9872568.

[6] I. Arpaci, S. Huang, M. Al-Emran, M.N. Al-Kabi, M. Peng, Predicting the COVID-19 infection with fourteen clinical features using machine learning classification algorithms, Multimed Tools Appl 80 (2021) 11943–11957. https://doi.org/10.1007/S11042-020-10340-7/FIGURES/5.

[7] M. Goldszmidt, Bayesian Network Classifiers, Wiley Encyclopedia of Operations Research and Management Science (2011). https://doi.org/10.1002/9780470400531.EORMS0099.

[8] Q. Wu, W. Zhao, Small-Cell Lung Cancer Detection Using a Supervised Machine Learning Algorithm, Proceedings - 2017 International Symposium on Computer Science and Intelligent Controls, ISCSIC 2017 2018-February (2017) 88–91. https://doi.org/10.1109/ISCSIC.2017.22.

[9] Y. Li, X. Wu, P. Yang, G. Jiang, Y. Luo, Machine Learning for Lung Cancer Diagnosis, Treatment, and Prognosis, Genomics Proteomics Bioinformatics 20 (2022) 850–866. https://doi.org/10.1016/J.GPB.2022.11.003.

[10] G.A.P. Singh, P.K. Gupta, Performance analysis of various machine learning-based approaches for detection and classification of lung cancer in humans, Neural Comput Appl 31 (2019) 6863–6877. https://doi.org/10.1007/S00521-018-3518-X/TABLES/8.

[11] N. Banerjee, S. Das, Prediction Lung Cancer- in Machine Learning Perspective, 2020 International Conference on Computer Science, Engineering and Applications, ICCSEA 2020 (2020). https://doi.org/10.1109/ICCSEA49143.2020.9132913.

[12] Z. Cai, D. Xu, Q. Zhang, J. Zhang, S.M. Ngai, J. Shao, Classification of lung cancer using ensemble-based feature selection and machine learning methods, Mol Biosyst 11 (2015) 791–800. https://doi.org/10.1039/C4MB00659C.

[13] S.S. Raoof, M.A. Jabbar, S.A. Fathima, Lung Cancer Prediction using Machine Learning: A Comprehensive Approach, 2nd International Conference on Innovative Mechanisms for Industry Applications, ICIMIA 2020 - Conference Proceedings (2020) 108–115. https://doi.org/10.1109/ICIMIA48430.2020.9074947.

[14] R. Hosseini Rad, S. Baniasadi, P. Yousefi, H. Morabbi Heravi, M. Shaban Al-Ani, M. Asghari Ilani, Presented a Framework of Computational Modeling to Identify the Patient Admission Scheduling Problem in the Healthcare System, J Healthc Eng 2022 (2022). https://doi.org/10.1155/2022/1938719.

[15] Y. Xie, W.Y. Meng, R.Z. Li, Y.W. Wang, X. Qian, C. Chan, Z.F. Yu, X.X. Fan, H.D. Pan, C. Xie, Q.B. Wu, P.Y. Yan, L. Liu, Y.J. Tang, X.J. Yao, M.F. Wang, E.L.H. Leung, Early lung cancer diagnostic biomarker discovery by machine learning methods, Transl Oncol 14 (2021) 100907. https://doi.org/10.1016/J.TRANON.2020.100907.

[16] P.R. Radhika, R.A.S. Nair, G. Veena, A Comparative Study of Lung Cancer Detection using Machine Learning Algorithms, Proceedings of 2019 3rd IEEE International Conference on Electrical, Computer and Communication Technologies, ICECCT 2019 (2019). https://doi.org/10.1109/ICECCT.2019.8869001.